\documentclass{article}

\PassOptionsToPackage{numbers,compress}{natbib}

\usepackage[preprint]{neurips_2026}

\usepackage[utf8]{inputenc} % allow utf-8 input
\usepackage[T1]{fontenc}    % use 8-bit T1 fonts
\usepackage{hyperref}       % hyperlinks
\usepackage{url}            % simple URL typesetting
\usepackage{booktabs}       % professional-quality tables
\usepackage{amsfonts}       % blackboard math symbols
\usepackage{amssymb}        % checkmark symbols
\usepackage{nicefrac}       % compact symbols for 1/2, etc.
\usepackage{microtype}      % microtypography
\usepackage{xcolor}         % colors
\usepackage[most]{tcolorbox} % boxed prompts

\usepackage{graphicx}  % include the graph
\usepackage{wrapfig}   % include side graph
\usepackage{placeins}  % keep appendix floats from crossing prompt boxes
\usepackage{colortbl, xcolor, booktabs, makecell, array, arydshln} % for table
\newcounter{algorithm}
\renewcommand{\thealgorithm}{\arabic{algorithm}}

\definecolor{nvgreen}{RGB}{118,185,0}
\hypersetup{colorlinks=true, urlcolor=nvgreen, linkcolor=nvgreen, citecolor=nvgreen}
\title{LongLive-RAG: A General Retrieval-Augmented Framework for Long Video Generation}

\author{%
  Qixin Hu\textsuperscript{1,2}\thanks{Work done during an internship at NVIDIA.} \quad
  Shuai Yang\textsuperscript{1} \quad
  Wei Huang\textsuperscript{1} \quad
  Song Han\textsuperscript{1,3} \quad
  Yukang Chen\textsuperscript{1} \\[4pt]
  \textsuperscript{1}NVIDIA \quad
  \textsuperscript{2}USC \quad
  \textsuperscript{3}MIT \\[10pt]
  \textbf{Project page:} \href{https://longlive-rag.github.io/}{\color{nvgreen}https://longlive-rag.github.io/}
}

\begin{document}

\maketitle

\begin{figure}[h]
    \centering
    \includegraphics[width=\linewidth]{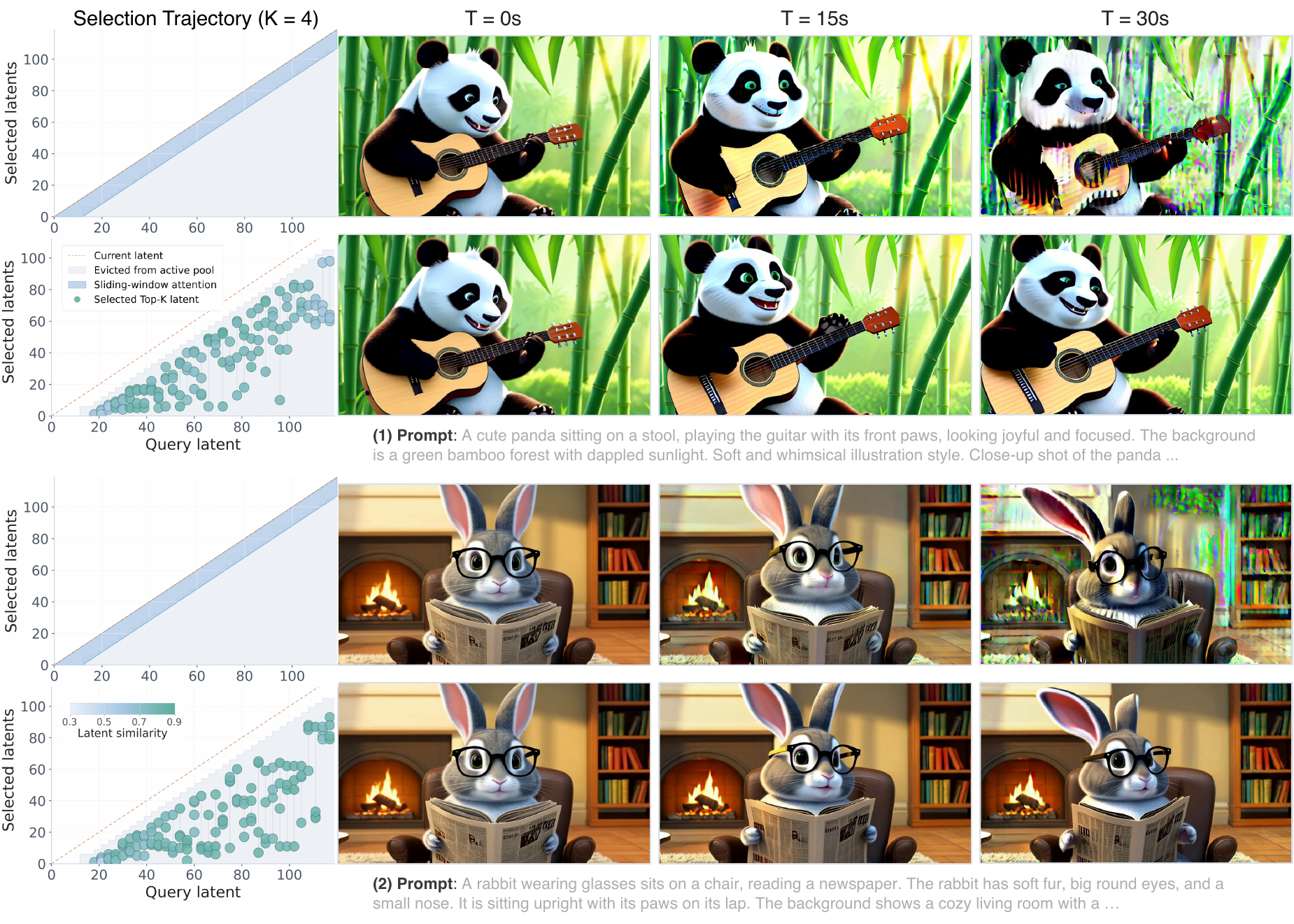}
    \caption{
    \textbf{LongLive-RAG} lets a video generator look back at useful parts of the video it has already generated.
    Left: for each new segment, LongLive-RAG actively searches the generated history and retrieves relevant past context.
    Right: using this retrieved context helps long videos keep subjects and scenes more stable and consistent over time.
    }
    \label{fig:teaser}
\end{figure}

\begin{abstract}
  Autoregressive (AR) video diffusion enables variable-length synthesis, but long-horizon generation often suffers from accumulated errors and identity drift. For efficiency, existing methods commonly adopt sliding-window attention during generation. This creates an irreversible generation trajectory: once the active window accumulates appearance errors, subsequent generations can only condition on this degraded trajectory and drift further away. We address this limitation by formulating long video generation as a retrieval-augmented generation (RAG) problem. Rather than relying solely on the recent window, we treat previously generated latents as a dynamic, searchable history. We propose \textbf{LongLive-RAG}, a general retrieval framework for AR video generation. At each new block, LongLive-RAG uses a query embedding to retrieve relevant historical latents. This lightweight retrieval step adds only a small overhead relative to generation and lets the generator condition on non-local context instead of only the recent window. To make retrieval more discriminative, we introduce the Window Temporal Delta Loss that suppresses redundant local similarity and encourages embeddings to capture meaningful temporal changes. Together, these components help reduce error accumulation caused by sliding-window attention. Experiments across multiple AR backbones and generation lengths show improved long-video quality and the best average VBench-Long rank. To our knowledge, among open-ended AR long video generation methods, LongLive-RAG is the first to formulate self-generated latent history as content-addressable retrieval memory. Code is available at \href{https://github.com/qixinhu11/LongLive-RAG}{\color{nvgreen}https://github.com/qixinhu11/LongLive-RAG}.
\end{abstract}

\section{Introduction}
\label{sec:introduction}

Recent video diffusion models can synthesize short clips with strong visual quality \citep{videogpt,magvit,magvitv2,videopoet,nova,pyramidalflow,wan,moviegen,kong2024hunyuanvideo,yang2024cogvideox}. Many applications, however, require longer videos in which subjects, backgrounds, and scene layouts remain consistent across tens or hundreds of seconds \citep{brooks2024video,feng2024matrix,hong2025relic,sun2025worldplay,yang2026stableworld,hu2023gaia,ren2025cosmos}. AR video diffusion is a practical formulation for this setting because it generates latent blocks causally and can continue beyond a fixed clip length \citep{diffusionforcing,causvid,selfforcing,causalforcing,rollingforcing,longlive,streamdit,streamingt2v}. In practice, long AR generation often suffers from error accumulation. To keep computation bounded, AR generators usually keep only the most recent blocks as context and discard earlier blocks. Once these recent blocks contain appearance drift, identity changes, or background artifacts, subsequent blocks may condition on these errors and further propagate them~\citep{liu2024sora,elmoghany2025survey,brooks2024video}.

Existing methods for long video generation address this problem by changing how history is kept. Attention-sink methods retain fixed early tokens or frames as anchors \citep{streamingllm,longlive,li2026rolling,lu2025reward}. Methods based on positional extrapolation extend the usable temporal range of the model \citep{rope,riflex,infinityrope}. Methods based on compressed-history tokens summarize older states into substitute tokens or recurrent memory \citep{deepforcing,yu2025videossm,kim2026memrope,zhang2025pretraining}. These methods can reduce error and improve stability, but they still have limitations. Fixed anchors may not match the current content, positional extrapolation does not prevent error accumulation once the visible context has drifted, and compressed-history tokens may lose critical native visual details for subsequent generations.

Our observation is that long video generation needs a way to use generated states when they are useful, instead of relying only on the recent window or a fixed summary of the past, as illustrated in Figure~\ref{fig:teaser}. The generated history may contain earlier context that better preserves the subject appearance, background layout, or scene structure needed by the current block. Retrieval provides a direct way to use this history. Before generating a new block, the model can search previously generated latents and bring back relevant historical context. When the recent window has drifted, this retrieved context can provide a reference and reduce dependence on the corrupted local context.

Based on this observation, we propose \textbf{LongLive-RAG}, a general retrieval-augmented framework for AR video generation. LongLive-RAG stores previously generated latents and builds compact embeddings for search. For each new block, an encoder maps the latest completed latent to a query embedding. The method retrieves the top-$K$ historical latents, combines them with the local latents, and uses them as the generator's attention context. The generator attends to the retrieved historical latents and local latents rather than compressed-history tokens. This gives the generator access to useful history outside the sliding window. It also keeps the base AR generator unchanged and adds only lightweight retrieval overhead relative to transformer attention.

The retrieval embedding must be suitable for search. Adjacent video latents are often very similar, so a reconstruction-only encoder can map many nearby blocks to almost identical embeddings. In that case, top-$K$ retrieval may return blocks that add little beyond the local window. We introduce a Window Temporal Delta Loss to reduce excessive similarity among nearby latents, and add a smoothing term to keep embeddings stable over time. Together, these losses make retrieval more discriminative while preserving visual information.

Our main contribution is to formulate open-ended AR video generation as retrieval over self-generated latents, together with a lightweight retrieval framework and an embedding objective that selects useful non-local context. LongLive-RAG retrieves useful past latents with a lightweight encoder and uses them as extra attention context. We compare it with two alternatives for addressing sliding-window limits: $\infty$-RoPE for positional extrapolation and Deep Forcing for compressed-history tokens. Across three AR backbones, Causal-Forcing, Self-Forcing, and LongLive, and three generation lengths, 30s, 60s, and 120s, LongLive-RAG achieves the best average VBench-Long rank and improves subject consistency, background consistency, motion smoothness, and imaging quality. Qualitative results and ablations further support these results.

\section{Related Work}
\label{sec:related_work}

This section briefly reviews related work; Appendix~\ref{app:related_work} gives the full discussion.

\paragraph{AR long video generation.}
AR video diffusion emits frames or latents causally for streaming and variable-length synthesis \citep{diffusionforcing,causvid,selfforcing,rollingforcing,longlive,streamdit,streamingt2v,magi1,skyreels,gu2025long,zhai2025stargen}.
Recent methods improve local denoising, self-generated context, or streaming attention \citep{causvid,selfforcing,selfforcingpp,causalforcing,contextforcing,rollingforcing,longlive}.
They mainly improve the quality or causality of each local rollout step; LongLive-RAG is orthogonal to these advances and keeps the generator unchanged.

\paragraph{Context visibility and memory.}
Long video generation systems expose history through sliding windows, fixed anchors, positional extrapolation, or compressed-history tokens \citep{streamingllm,longlive,li2026rolling,lu2025reward,rope,riflex,infinityrope,deepforcing,contextforcing,yu2025videossm,kim2026memrope,zhang2025pretraining,zhang2025packing}.
These strategies differ in how much history is kept and how it is stored; LongLive-RAG instead searches the generated history and brings back context for attention.

\paragraph{Retrieval memory in video generation.}
Retrieval is natural in world models when geometry, camera pose, or scene coordinates are available \citep{hu2023gaia,feng2024matrix,li2025vmem,yu2025context,xiao2025worldmem,sun2025worldplay,yang2026stableworld,wu2026infinite}.
Open-ended text-to-video generation does not provide such explicit retrieval cues by default, so useful history must be found from the generated content itself; LongLive-RAG searches the generated history and uses the matched context during rollout.

\begin{figure}
    \centering
    \includegraphics[width=\linewidth]{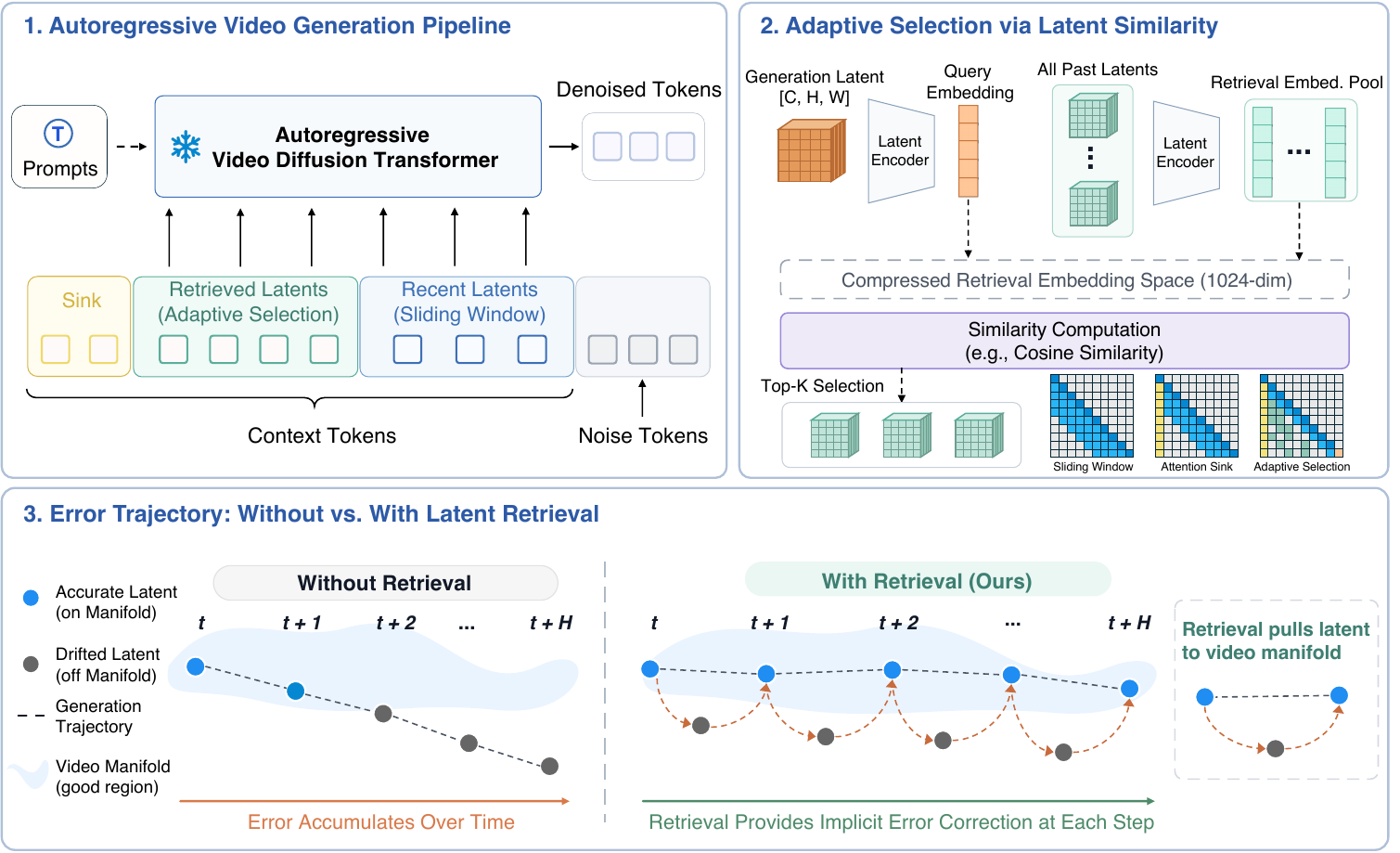}
    \caption{
    Overview of LongLive-RAG.
    (1) The AR video diffusion transformer produces a generated latent and assembles context tokens.
    (2) A latent encoder maps the latent to a compact query embedding and searches the retrieval pool.
    (3) Retrieved latents provide context; when drift enters the trajectory and errors accumulate over time, retrieval provides implicit correction at each step.
    }
    \label{fig:method}
\end{figure}

\section{Method}
\label{sec:method}

LongLive-RAG augments AR video diffusion with retrieval over self-generated latents while keeping the base generator fixed.
Section~\ref{subsec:sliding_window_inference} defines the sliding-window and LongLive-RAG context assembly rules.
Section~\ref{subsec:latent_history_bank} describes the paired embedding/context history banks and the retrieval rule.
Section~\ref{subsec:temporal_constraints} explains retrieval embedding space training.
Section~\ref{subsec:retrieval_inference} summarizes the inference.

\subsection{AR Context Assembly}
\label{subsec:sliding_window_inference}

\paragraph{Sliding-window context.}
At block $t$, an AR video diffusion model denoises the current latent representation while attending to context from previously completed blocks \citep{causvid,selfforcing,rollingforcing,longlive}. Standard inference typically keeps a bounded sliding-window context
\begin{equation}
\mathcal{A}^{\mathrm{sw}}_t =
[\mathcal{C}_{\mathrm{sink}}\Vert \mathcal{C}_{\mathrm{loc}}],
\label{eq:sliding_window_context}
\end{equation}
where $\mathcal{C}_{\mathrm{sink}}$ stores optional fixed early context and $\mathcal{C}_{\mathrm{loc}}$ stores context from recent completed blocks \citep{streamingllm,longlive,yu2025videossm}.
Once earlier context is dropped, the generator cannot use it again; if the recent context has drifted, the next block is still generated from this drifted context.

\paragraph{LongLive-RAG context.}
LongLive-RAG does not update the base generator or add trainable layers inside the denoising backbone.
Instead, it uses an offline-trained retrieval encoder and adds relevant historical context entries directly to the attention context:
\begin{equation}
\mathcal{A}^{\mathrm{rag}}_t =
[\mathcal{C}_{\mathrm{sink}}\Vert \mathcal{M}_t \Vert \mathcal{C}_{\mathrm{loc}}].
\label{eq:longliverag_context}
\end{equation}
Here $\mathcal{M}_t$ denotes the retrieved non-local context entries for block $t$. The current block is then denoised with the retrieved history and local context. Importantly, LongLive-RAG does not add attention layers or change the denoising rule.

\subsection{Indexing the History}
\label{subsec:latent_history_bank}

\paragraph{Compact embeddings for search.}
Let $\hat{x}_t \in \mathbb{R}^{C \times H \times W}$ be the clean latent at block $t$.
This latent remains spatially dense because the generator needs it to preserve visual quality and fine details during denoising, as in WAN~\citep{wan}.
Retrieval has a different requirement: it only needs a discriminative key for finding useful history, not the full representation used for synthesis.
LongLive-RAG therefore performs search in a compressed retrieval embedding space instead of directly comparing all historical latents with spatial shape $[C,H,W]$.
For a completed block $i$, an encoder produces a 1024-dimensional embedding $v_i$, and the base backbone provides the corresponding context entry $\mathcal{C}_i$.
The historical search bank is $\mathcal{H}_v=\{(i,v_i)\}$, paired with a context bank $\mathcal{H}_{\mathrm{ctx}}=\{(i,\mathcal{C}_i)\}$.
The embeddings are used for retrieval, while the generator attends to the matched context.

We train a latent autoencoder with encoder $E_\psi$ and decoder $D_\psi$ using reconstruction loss,
\begin{equation}
\mathcal{L}_{\mathrm{rec}} =
\|D_\psi(E_\psi(\hat{x}_t)) - \hat{x}_t\|_2^2,
\label{eq:rec_loss}
\end{equation}
and use $v_t=E_\psi(\hat{x}_t)$ as the retrieval embedding.
Here $E_\psi$ is the encoder and $v_t \in \mathbb{R}^{1024}$ lies in the compact search space.
The encoder is trained offline on clean latents with the base generator fixed; the decoder is used only to shape this space and is not inserted into the generator at inference.
After block $t$ is completed, LongLive-RAG computes $v_t$ and associates it with the block context $\mathcal{C}_t$.
Entries remain in the rolling local cache while they are recent; when they leave the local window, their paired $(v_i,\mathcal{C}_i)$ entries are added to $\mathcal{H}_v$ and $\mathcal{H}_{\mathrm{ctx}}$.

\paragraph{Similarity retrieval.}
For block $t$, LongLive-RAG performs retrieval once before denoising and reuses the selected context for all $N$ denoising steps.
The query is the embedding of the most recent completed latent, i.e., $v_q=v_{t-1}$ after the previous block has been finalized.
LongLive-RAG ranks stored embeddings with cosine similarity,
\begin{equation}
a_i = \cos(v_q, v_i), \qquad i \in \mathcal{P}_t,
\label{eq:retrieval_score}
\end{equation}
where $\mathcal{P}_t$ skips the most recent history entries to avoid retrieving near-duplicate recent context.
The top-$K$ indices $I_t=\mathrm{Top}\text{-}K(\{a_i\}_{i\in\mathcal{P}_t})$ select matched entries from $\mathcal{H}_{\mathrm{ctx}}$:
\begin{equation}
\mathcal{M}_t = [\mathcal{C}_i]_{i\in I_t}.
\label{eq:retrieved_context}
\end{equation}
These retrieved context entries are combined with the local cache in the current attention context.

\paragraph{Context handling.}
Each entry in $\mathcal{H}_{\mathrm{ctx}}$ follows the same interface used by the local window.
LongLive-RAG does not introduce new attention layers; it only changes which historical context entries are exposed to attention at each block.
Retrieved entries are therefore handled in the same way as local context entries inside the original generator.

\paragraph{Retrieval overhead.}
\begin{wraptable}{r}{0.47\textwidth}
    \vspace{-1.0em}
    \centering
    \caption{
    120s retrieval overhead.
    }
    \label{tab:system_overhead}
    \small
    \setlength{\tabcolsep}{3.5pt}
    \begin{tabular}{@{}lcc@{}}
        \toprule
        Component & ms/block & Total (ms) \\
        \midrule
        Latent encoding & 3.96 & 480 \\
        Top-$K$ search & 0.08 & 10 \\
        \midrule
        Total & 4.08 & 490 \\
        \bottomrule
    \end{tabular}
    \vspace{-1.0em}
\end{wraptable}
Table~\ref{tab:system_overhead} reports the additional runtime introduced by LongLive-RAG retrieval over a 120s rollout.
LongLive-RAG performs retrieval once per AR block and adds 4.08 ms per block, totaling 490 ms of retrieval overhead: 480 ms from latent encoding and 10 ms from top-$K$ search.
The overhead is dominated by encoding; similarity search is negligible at this scale.
For long video generation, this cost is small compared with the diffusion rollout itself.

\subsection{Learning the Embedding Space}
\label{subsec:temporal_constraints}

\begin{figure}[t]
    \centering
    \includegraphics[width=\linewidth]{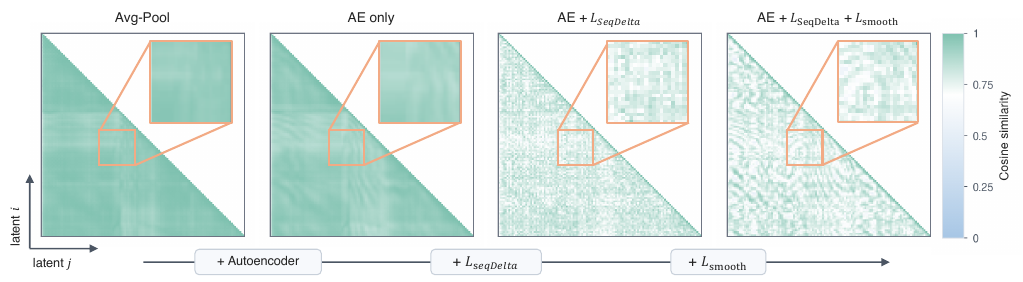}
    \caption{
    Embedding-space analysis.
    Darker green indicates a higher cosine similarity between the embeddings. Relying on reconstruction alone preserves content but leaves temporally nearby latents overly similar; the temporal delta separates redundant local states, and smoothing effectively stabilizes the embedding trajectory along the entire generation trajectory.
    }
    \label{fig:analysis}
\end{figure}

\paragraph{Why reconstruction is not enough.}
The search space has a different requirement from the VAE latent space used by video generators such as WAN~\citep{wan}.
For synthesis, adjacent latents should change smoothly and preserve dense visual detail.
For retrieval, however, the embedding must also decide which history is worth bringing back into attention.
If the embedding only optimizes reconstruction, it tends to preserve the local continuity of video too well: nearby latents become almost interchangeable search keys.
Figure~\ref{fig:analysis} illustrates this behavior.
With AE-only training, the similarity map contains a broad high-similarity band around the diagonal, meaning that many neighboring blocks look equally valid to the retriever.
In inference, such an index can waste the top-$K$ budget on locally redundant states, adding little beyond the recent cache.

This motivates an anti-collapse term at the temporal scale where redundancy is most common.
We do not want to repel all frames from each other, since the same subject or scene may legitimately reappear later.
Instead, we only discourage excessive similarity within a short window, where frames are likely to be visually redundant and already covered by the local cache.
The loss is therefore local and margin-based: pairs below the margin are not pushed apart, and long-range repetitions are not treated as negatives.
We first define a pairwise temporal delta penalty,
\begin{equation}
\mathcal{L}_{\Delta}(v_t, v_{\mathrm{ref}}) =
\lambda_{\Delta}\max(0,\, \cos(v_t, v_{\mathrm{ref}}) - m),
\label{eq:delta_loss}
\end{equation}
which penalizes local pairs whose cosine similarity exceeds a margin $m$.
The Window Temporal Delta Loss averages this penalty over a local temporal window.
For a sequence of length $T$ and window size $w$, let $\bar{w}=\min(w,T-1)$.
\begin{equation}
\mathcal{L}_{\mathrm{SeqDelta}} =
\frac{1}{\bar{w}}\sum_{\tau=1}^{\bar{w}}
\frac{1}{T-\tau}\sum_{t=\tau+1}^{T}
\mathcal{L}_{\Delta}(v_t, v_{t-\tau}).
\label{eq:seq_delta_loss}
\end{equation}

\paragraph{Stable embedding trajectory.}
Temporal separation alone is still insufficient.
If the embedding moves too sharply from one block to the next, the top-$K$ set can change for reasons unrelated to meaningful content changes, which may make the retrieved context unstable.
The right behavior is visible in Figure~\ref{fig:analysis}: after local redundancy is reduced, the embedding trajectory should remain organized rather than noisy.
We therefore add a second-order smoothness penalty
\begin{equation}
\mathcal{L}_{\mathrm{Smooth}} =
\lambda_{\mathrm{smooth}} \frac{1}{T-2}
\sum_{t=3}^{T} \| v_t - 2v_{t-1} + v_{t-2} \|_2,
\label{eq:smooth_loss}
\end{equation}
Combining these terms gives the final training objective:
\begin{equation}
\mathcal{L}_{\mathrm{Total}} =
\mathcal{L}_{\mathrm{rec}}
+ \mathcal{L}_{\mathrm{SeqDelta}}
+ \mathcal{L}_{\mathrm{Smooth}}.
\label{eq:total_loss}
\end{equation}
The three terms play distinct roles.
Reconstruction keeps the embedding tied to visual content; temporal delta makes nearby redundant states less likely to dominate top-$K$ retrieval; and smoothing keeps the embedding trajectory stable across the generation trajectory.
This is the design principle behind Figure~\ref{fig:analysis}: LongLive-RAG does not seek a generic compressed representation, but a search geometry that matches the needs of non-local context selection.

\subsection{Inference}
\label{subsec:retrieval_inference}

% LongLive-RAG inference algorithm (wrapfigure on the left).
% Required packages: xcolor, wrapfig, booktabs (already loaded by main.tex).
% Required counter: algorithm (declared in main.tex).

\definecolor{algcomment}{gray}{0.45}
\providecommand{\algcmt}[1]{{\color{algcomment}\itshape #1}}

\begin{wrapfigure}[26]{l}{0.56\textwidth}
\vspace{-0.4em}
\refstepcounter{algorithm}\label{alg:longliverag_inference}
\begin{minipage}{0.56\textwidth}
\small
\setlength{\tabcolsep}{1.5pt}
\renewcommand{\arraystretch}{0.92}
\begin{tabular}{@{}r@{\hspace{0.35em}}p{0.88\linewidth}@{}}
\toprule
\multicolumn{2}{@{}p{\linewidth}@{}}{\textbf{Algorithm \thealgorithm\ \ LongLive-RAG inference}}\\
\midrule
\multicolumn{2}{@{}p{\linewidth}@{}}{\textbf{Require:} $G_\theta, E_\psi$; $T, N, L, K, R$}\\
\multicolumn{2}{@{}p{\linewidth}@{}}{\textbf{State:} $\mathcal{C}_{\mathrm{sink}}, \mathcal{C}_{\mathrm{loc}}, \mathcal{H}_{\mathrm{ctx}}, \mathcal{H}_v$}\\
\midrule
1 & $\mathcal{C}_{\mathrm{sink}}, \mathcal{C}_{\mathrm{loc}}, \mathcal{H}_{\mathrm{ctx}}, \mathcal{H}_v \leftarrow \emptyset$\\
2 & \textbf{for} block $t = 1, \ldots, T$ \textbf{do}\\
3 & \quad \algcmt{// Step 1: query searches pool}\\
4 & \quad $I_t \leftarrow \emptyset$\\
5 & \quad \textbf{if} $|\mathcal{H}_v| > R$ \textbf{then}\\
6 & \quad\quad $v_q \leftarrow v_{t-1}$ \hfill \algcmt{$\triangleright$ latest embedding}\\
7 & \quad\quad $I_t \leftarrow$ top-$K$ matches in $\mathcal{H}_v$ outside recent $R$\\
8 & \quad \textbf{end if}\\
9 & \quad \algcmt{// Step 2: denoise latent $t$}\\
10 & \quad $\mathcal{M}_t \leftarrow [\mathcal{H}_{\mathrm{ctx}}[i]]_{i\in I_t}$\\
11 & \quad $\mathcal{A}_t \leftarrow [\mathcal{C}_{\mathrm{sink}} \Vert \mathcal{M}_t \Vert \mathcal{C}_{\mathrm{loc}}]$\\
12 & \quad \textbf{for} step $s = 1, \ldots, N$ \textbf{do}\\
13 & \quad\quad $x_t^{s+1} \leftarrow G_\theta(x_t^s;\, \mathcal{A}_t)$\\
14 & \quad \textbf{end for}\\
15 & \quad \algcmt{// Step 3: encoder updates pool}\\
16 & \quad $\hat{x}_t \leftarrow x_t^{N+1}$;\ $v_t \leftarrow E_\psi(\hat{x}_t)$\\
17 & \quad Append context and $v_t$ to $\mathcal{C}_{\mathrm{loc}}$\\
18 & \quad \textbf{if} $|\mathcal{C}_{\mathrm{loc}}| > L$ \textbf{then}\\
19 & \quad\quad Add oldest entry to $(\mathcal{H}_{\mathrm{ctx}}, \mathcal{H}_v)$\\
20 & \quad\quad Keep the latest $L$ entries in $\mathcal{C}_{\mathrm{loc}}$\\
21 & \quad \textbf{end if}\\
22 & \textbf{end for}\\
\bottomrule
\end{tabular}
\end{minipage}
\end{wrapfigure}

Algorithm~\ref{alg:longliverag_inference} gives the inference procedure with a reduced notation set.
Here $G_\theta$ is the frozen AR generator and $E_\psi$ is the encoder.
$T$ is the number of latents, $N$ is the number of denoising steps per block, $L$ is the local-window size, $K$ is the retrieval budget, and $R$ is the recency guard that prevents the search from selecting near-duplicate recent context.
The state consists of an optional sink cache $\mathcal{C}_{\mathrm{sink}}$, a rolling local cache $\mathcal{C}_{\mathrm{loc}}$ for the latest context, the pool $\mathcal{H}_v$, and a paired historical context pool $\mathcal{H}_{\mathrm{ctx}}$.
Entry $i$ of $\mathcal{H}_v$ is the compact search key for entry $i$ of $\mathcal{H}_{\mathrm{ctx}}$.
At block $t$, $v_q=v_{t-1}$ is the query from the latest completed latent, $I_t$ is the selected top-$K$ history indices, $\mathcal{M}_t$ is the matched context, and $\mathcal{A}_t=[\mathcal{C}_{\mathrm{sink}}\Vert\mathcal{M}_t\Vert\mathcal{C}_{\mathrm{loc}}]$ is the attention context.
When no eligible historical entry exists, $I_t$ is empty and the method reduces to the base sink-plus-local context.
The latent $x_t^s$ denotes block $t$ at denoising step $s$, $\hat{x}_t=x_t^{N+1}$ is the clean completed latent, and $v_t=E_\psi(\hat{x}_t)$ is the embedding stored with the completed local entry before later offloading to the historical pool.
The notation $|\cdot|$ denotes the number of entries in a bank, and $\Vert$ denotes concatenation along the context dimension.
As illustrated in Figure~\ref{fig:method}(3), this gives later blocks access to content-relevant non-local context while retaining the original AR generator.

\begin{figure}[h]
    \centering
    \includegraphics[width=\linewidth]{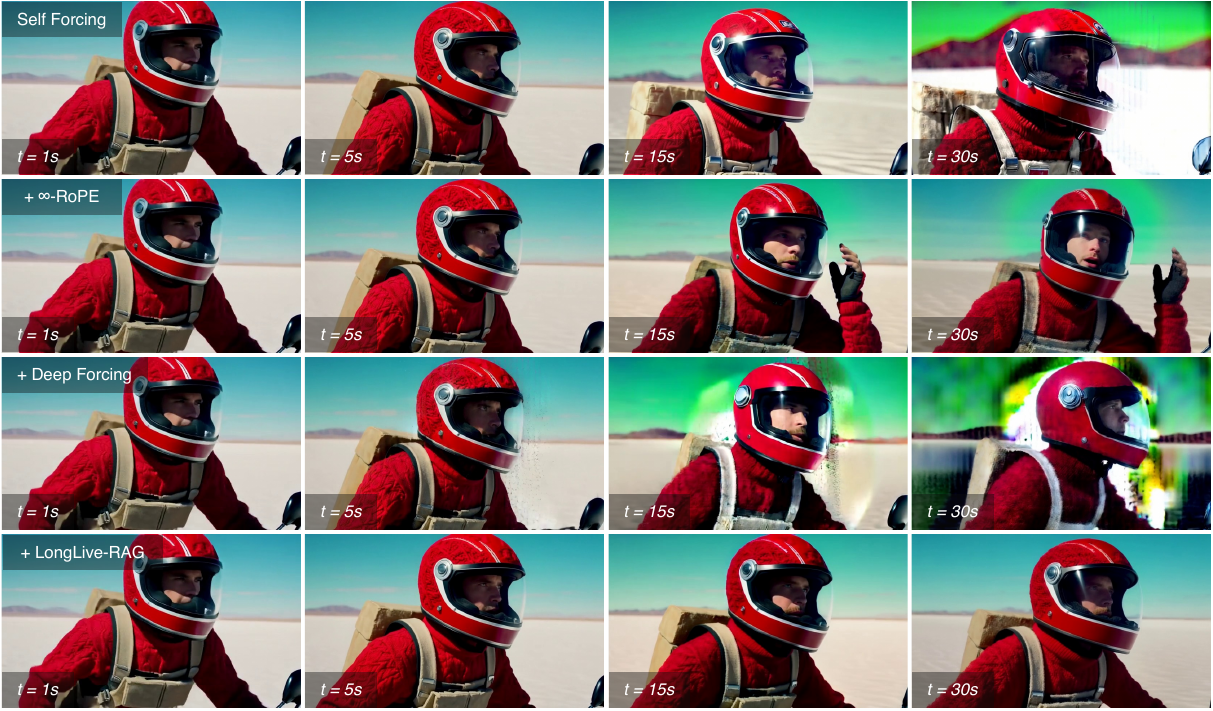}
    \includegraphics[width=\linewidth]{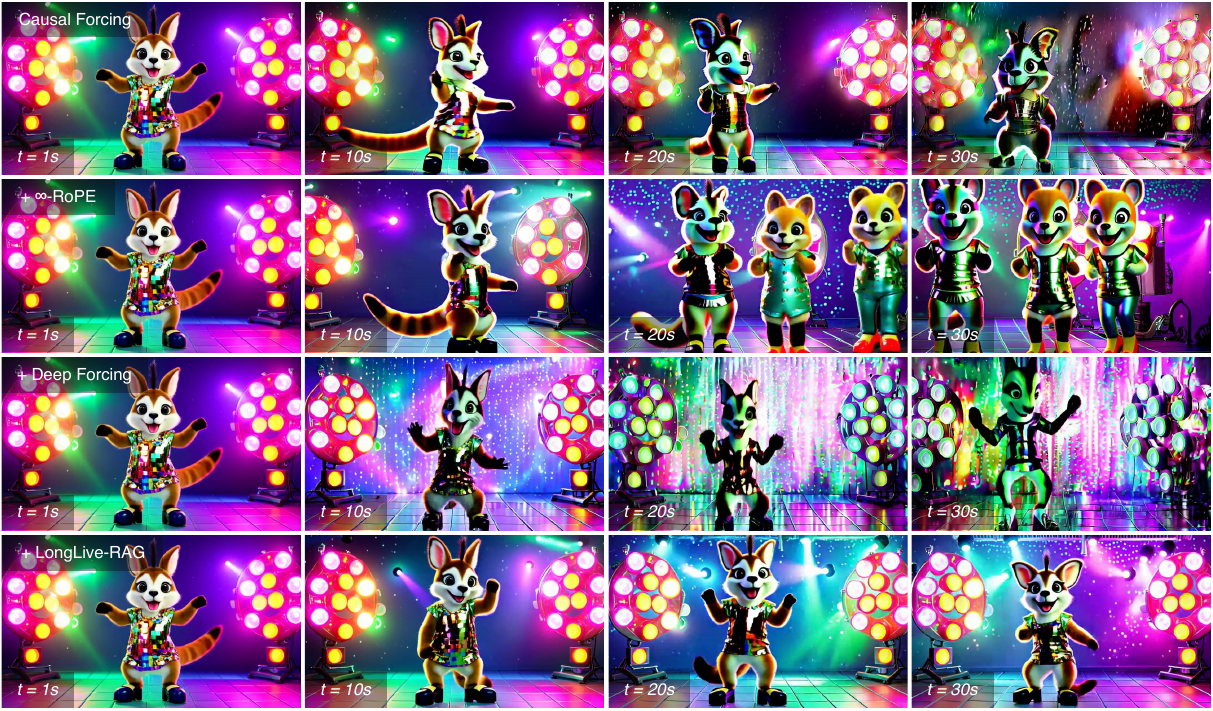}
    \caption{
    Qualitative comparison on 30s generations.
    Rows compare the base model, $\infty$-RoPE, Deep Forcing, and LongLive-RAG; columns show later timestamps.
    The displayed baselines show color shifts, duplicated subjects, or background artifacts.
    }
    \label{fig:compare}
\end{figure}

\section{Experiments}
\label{sec:experiments}

We evaluate whether LongLive-RAG improves long-video quality across three AR backbones and three generation lengths, 30s, 60s, and 120s, and whether learned retrieval contributes to the gains.

\subsection{Setup}
\label{subsec:exp_setup}

\paragraph{Implementation details.}
We evaluate LongLive-RAG on three AR backbones: Causal-Forcing~\citep{causalforcing}, Self-Forcing~\citep{selfforcing}, and LongLive~\citep{longlive}.
For each backbone, we compare the base model, $\infty$-RoPE~\citep{infinityrope}, Deep Forcing~\citep{deepforcing}, and LongLive-RAG.
$\infty$-RoPE represents positional extrapolation, while Deep Forcing represents compressed-history tokens.
All inference runs use the same base sampling settings, sink size $1$, and total attention window size $12$.
LongLive-RAG uses $K=6$, giving the context layout $[1\ \mathrm{sink}\Vert 6\ \mathrm{retrieved\ context}\Vert 5\ \mathrm{local\ window}]$.
Under this fixed context budget, the compared methods differ only in how the available slots are filled: the base model and $\infty$-RoPE use sink-plus-local context, while Deep Forcing uses compressed-history context following its inference rule.
The reported inference and evaluation runs for the quantitative tables take about one week on 6 NVIDIA RTX A6000 GPUs, excluding retrieval-encoder data construction and autoencoder training.
More implementation details are provided in Appendix~\ref{app:experimental_details}.

\paragraph{Evaluation protocol.}
We use all 128 prompts from MovieGenBench~\citep{moviegen}.
Following Self-Forcing~\citep{selfforcing}, all prompts are refined using Qwen2.5-7B-Instruct~\citep{qwen25}.
Table~\ref{tab:main} reports VBench-Long metrics~\citep{vbench} for 30s, 60s, and 120s generations.
Table~\ref{tab:ablation_embedding} reports ablations on 30s Causal-Forcing and auxiliary VLM scores.
The VLM prompt is provided in Appendix~\ref{app:vlm_eval}.

% Required packages (add to preamble):
% \usepackage{colortbl, xcolor, booktabs, makecell, array, arydshln}

% Sparse grayscale table theme.
\definecolor{headerbg}{HTML}{EEEEEE}
\definecolor{basebg}{HTML}{F8F8F8}

\newcommand{\first}[1]{\textbf{#1}}
\newcommand{\second}[1]{\underline{#1}}

\begin{table*}[t]
\centering
\caption{
VBench-Long results for 30s, 60s, and 120s generation.
Each block fixes a base model and compares it with $\infty$-RoPE~\citep{infinityrope}, Deep Forcing~\citep{deepforcing}, and LongLive-RAG.
Bold/underline mark best/second-best values; Avg. Rank is averaged over the six metrics.
Avg. Rank is computed from unrounded metric values.
}
\label{tab:main}
\scriptsize
\renewcommand{\arraystretch}{1.08}
\setlength{\tabcolsep}{4pt}
\resizebox{\textwidth}{!}{%
\begin{tabular}{l cccccc c}
\toprule
\rowcolor{headerbg}
Method
& \makecell{Subject\\Consistency} $\uparrow$
& \makecell{Background\\Consistency} $\uparrow$
& \makecell{Motion\\Smoothness} $\uparrow$
& \makecell{Dynamic\\Degree} $\uparrow$
& \makecell{Aesthetic\\Quality} $\uparrow$
& \makecell{Imaging\\Quality} $\uparrow$
& \makecell{Avg.\\Rank} $\downarrow$ \\
\midrule
\multicolumn{8}{c}{\textit{30s generation}} \\
\midrule
\rowcolor{basebg}
Self-Forcing~\citep{selfforcing}    & 96.21           & 95.39           & 98.39           & \first{52.03}   & 56.69           & 63.31           & 3.17 \\
+ $\infty$-RoPE~\citep{infinityrope} & \second{97.32} & \second{96.38} & \second{98.59} & \second{46.82} & \second{56.78} & \second{63.93} & \second{2.00} \\
+ Deep Forcing~\citep{deepforcing} & 97.04           & 96.02           & 98.57           & 38.85           & 56.44           & 61.91           & 3.50 \\
\textbf{+ LongLive-RAG (Ours)}        & \first{97.57}  & \first{96.56}  & \first{98.76}  & 42.24           & \first{57.17}  & \first{65.43}  & \first{1.33} \\
\cmidrule(lr){1-8}
\rowcolor{basebg}
LongLive~\citep{longlive}              & 97.35           & 96.15           & 98.70           & 44.74           & \first{59.38}  & \second{68.15} & 2.67 \\
+ $\infty$-RoPE~\citep{infinityrope} & 97.27           & 96.19           & 98.68           & \first{48.18}   & 58.69           & 67.99           & 3.17 \\
+ Deep Forcing~\citep{deepforcing} & \second{97.52} & \first{96.43}  & \first{98.82}  & 41.46           & 59.00           & 67.61           & \second{2.50} \\
\textbf{+ LongLive-RAG (Ours)}        & \first{97.53}  & \second{96.39} & \second{98.77} & \second{44.84} & \second{59.24} & \first{68.42}  & \first{1.67} \\
\cmidrule(lr){1-8}
\rowcolor{basebg}
Causal-Forcing~\citep{causalforcing}         & \second{94.60} & \second{94.68} & \second{96.56} & 73.96           & 54.58           & 65.53           & 3.00 \\
+ $\infty$-RoPE~\citep{infinityrope} & 93.93           & 94.11           & 96.21           & \first{90.83}   & \second{55.42} & \second{68.26} & \second{2.33} \\
+ Deep Forcing~\citep{deepforcing} & 93.52           & 93.86           & 95.84           & \second{84.79} & 55.03           & 66.07           & 3.33 \\
\textbf{+ LongLive-RAG (Ours)}        & \first{95.43}  & \first{94.79}  & \first{97.16}  & 82.29           & \first{57.31}  & \first{70.07}  & \first{1.33} \\
\addlinespace[3pt]
\midrule
\multicolumn{8}{c}{\textit{60s generation}} \\
\midrule
\rowcolor{basebg}
Self-Forcing~\citep{selfforcing}    & 95.84           & 95.27           & 98.20           & \first{51.72}   & 56.05           & 62.22           & 3.33 \\
+ $\infty$-RoPE~\citep{infinityrope} & \second{97.24} & \second{96.24} & \second{98.58} & \second{46.64} & 56.09           & \second{63.28} & \second{2.17} \\
+ Deep Forcing~\citep{deepforcing} & 96.08           & 95.38           & 98.24           & 41.44           & \second{56.68} & 60.81           & 3.17 \\
\textbf{+ LongLive-RAG (Ours)}        & \first{97.60}  & \first{96.51}  & \first{98.70}  & 44.69           & \first{57.19}  & \first{64.97}  & \first{1.33} \\
\cmidrule(lr){1-8}
\rowcolor{basebg}
LongLive~\citep{longlive}              & 97.13           & 95.89           & 98.61           & 44.56           & \second{58.17} & \second{67.56} & 2.83 \\
+ $\infty$-RoPE~\citep{infinityrope} & 97.00           & 95.85           & 98.53           & \first{53.36}   & 57.48           & 66.94           & 3.33 \\
+ Deep Forcing~\citep{deepforcing} & \second{97.17} & \second{96.04} & \first{98.73}  & 45.13           & 57.48           & 67.27           & \second{2.50} \\
\textbf{+ LongLive-RAG (Ours)}        & \first{97.32}  & \first{96.08}  & \second{98.62} & \second{49.90} & \first{58.30}  & \first{67.79}  & \first{1.33} \\
\cmidrule(lr){1-8}
\rowcolor{basebg}
Causal-Forcing~\citep{causalforcing}         & 93.52           & 94.12           & 95.74           & 72.32           & 51.24           & 62.30           & 3.83 \\
+ $\infty$-RoPE~\citep{infinityrope} & 93.81           & 93.78           & 96.09           & \first{92.47}   & \second{54.42} & \second{67.50} & 2.50 \\
+ Deep Forcing~\citep{deepforcing} & \second{94.27} & \second{94.18} & \first{96.62}  & 78.59           & 52.12           & 64.25           & \second{2.33} \\
\textbf{+ LongLive-RAG (Ours)}        & \first{94.29}  & \first{94.24}  & \second{96.48} & \second{88.20} & \first{54.95}  & \first{68.16}  & \first{1.33} \\
\addlinespace[3pt]
\midrule
\multicolumn{8}{c}{\textit{120s generation}} \\
\midrule
\rowcolor{basebg}
Self-Forcing~\citep{selfforcing}    & 96.12           & 95.32           & 98.27           & 43.39           & \second{55.64} & 61.57           & 3.33 \\
+ $\infty$-RoPE~\citep{infinityrope} & \second{97.15} & 96.09           & 98.55           & \first{46.29}   & 55.11           & \second{61.81} & \second{2.33} \\
+ Deep Forcing~\citep{deepforcing} & 96.92           & \first{96.83}  & \first{98.97}  & 15.23           & 52.84           & 57.93           & 2.83 \\
\textbf{+ LongLive-RAG (Ours)}        & \first{97.64}  & \second{96.40} & \second{98.75} & \second{44.10} & \first{56.30}  & \first{64.16}  & \first{1.50} \\
\cmidrule(lr){1-8}
\rowcolor{basebg}
LongLive~\citep{longlive}              & 96.93           & 95.64           & \second{98.58} & 47.12           & \first{57.90}  & \second{66.95} & \second{2.50} \\
+ $\infty$-RoPE~\citep{infinityrope} & 96.81           & 95.65           & 98.48           & \first{53.59}   & 56.73           & 66.19           & 3.17 \\
+ Deep Forcing~\citep{deepforcing} & \second{97.17} & \second{95.72} & 98.57           & 46.03           & 56.98           & 66.03           & 3.00 \\
\textbf{+ LongLive-RAG (Ours)}        & \first{97.22}  & \first{95.88}  & \first{98.62}  & \second{50.25} & \second{57.57} & \first{66.95}  & \first{1.33} \\
\cmidrule(lr){1-8}
\rowcolor{basebg}
Causal-Forcing~\citep{causalforcing}         & 92.98           & \first{94.66}  & 95.41           & 63.79           & 47.31           & 58.23           & 3.33 \\
+ $\infty$-RoPE~\citep{infinityrope} & \second{93.45} & 93.56           & \second{95.95} & \first{93.44}   & \second{53.47} & \second{66.81} & \second{2.17} \\
+ Deep Forcing~\citep{deepforcing} & 93.19           & 94.04           & 95.60           & 76.45           & 46.86           & 61.58           & 3.17 \\
\textbf{+ LongLive-RAG (Ours)}        & \first{94.38}  & \second{94.08} & \first{96.56}  & \second{90.21} & \first{54.82}  & \first{68.23}  & \first{1.33} \\
\bottomrule
\end{tabular}%
}
\end{table*}

\subsection{Results}
\label{subsec:main_results}

\paragraph{Qualitative results.}
Figure~\ref{fig:compare} compares LongLive-RAG with alternative history-handling methods, including $\infty$-RoPE and Deep Forcing, on 30s rollouts.
LongLive-RAG better preserves subject and background appearance, while other methods can show appearance shifts, duplicated subjects, or color artifacts.
These examples support the same trend measured by Table~\ref{tab:main}: retrieving selected non-local latents provides useful context beyond the fixed local window or compressed history.

\paragraph{Quantitative results.}
Table~\ref{tab:main} shows that LongLive-RAG obtains the lowest Avg. Rank across all base-model and duration blocks.
The gains are consistent across 30s, 60s, and 120s rollouts, suggesting that retrieval remains useful as generation length increases.
Improvements are most visible on quality and consistency metrics, including subject consistency, background consistency, motion smoothness, and imaging quality.
These results indicate that selected non-local latents help the generator preserve appearance and scene structure beyond the recent window.

\paragraph{Auxiliary VLM evaluation.}
Table~\ref{tab:vlm_eval} also reports VLM exposure scores on 30s generations as an auxiliary check.
The scores show the same overall trend as the VBench-Long results.

\subsection{Ablations}
\label{subsec:ablations}

% Ablation table and retrieval-budget plots.

\begin{table*}[t]
\centering
\scriptsize
\begin{minipage}[t]{0.445\textwidth}
\vspace{0pt}
\centering
\textbf{(a) Embedding space.}\\[0.10em]
\renewcommand{\arraystretch}{1.25}
\setlength{\tabcolsep}{3pt}
\begin{tabular}{lcccc}
\toprule
\rowcolor{headerbg}
Method
& \makecell{Subj.\\Cons.} $\uparrow$
& \makecell{Bg.\\Cons.} $\uparrow$
& \makecell{Motion\\Smooth.} $\uparrow$
& \makecell{Imaging\\Quality} $\uparrow$ \\
\midrule
Random retrieval & 94.54 & 94.32 & 96.81 & 68.79 \\
Avg-pool desc. & 94.77 & 94.49 & 96.76 & 69.11 \\
AE only & \underline{94.82} & 94.49 & 96.87 & \underline{69.48} \\
AE + SeqDelta & 94.76 & \underline{94.54} & \underline{97.04} & 69.14 \\
\rowcolor{basebg}
\textbf{Ours} & \textbf{95.43} & \textbf{94.79} & \textbf{97.16} & \textbf{70.07} \\
\bottomrule
\end{tabular}

\vspace{1.10em}
\textbf{(b) Auxiliary VLM evaluation. (Max: 5)}\\[0.10em]
\renewcommand{\arraystretch}{1.25}
\setlength{\tabcolsep}{5pt}
\begin{tabular}{lccc}
\toprule
\rowcolor{headerbg}
Variant & Causal Forc. & Self Forc. & LongLive \\
\midrule
Base & 2.60 & 3.50 & 4.65 \\
+ $\infty$-RoPE & 4.10 & 4.15 & 4.35 \\
+ Deep Forcing & 3.55 & 4.35 & 4.70 \\
\rowcolor{basebg}
\textbf{+ LongLive-RAG} & \textbf{4.70} & \textbf{4.45} & \textbf{4.75} \\
\bottomrule
\end{tabular}
\end{minipage}
\hfill
\begin{minipage}[t]{0.535\textwidth}
\vspace{1.4em}
\centering
\textbf{(c) Retrieval budget.}\\[0.10em]
\setlength{\tabcolsep}{1pt}
\begin{tabular}{@{}cc@{}}
\includegraphics[width=0.46\linewidth]{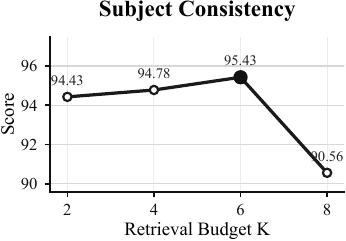} &
\includegraphics[width=0.46\linewidth]{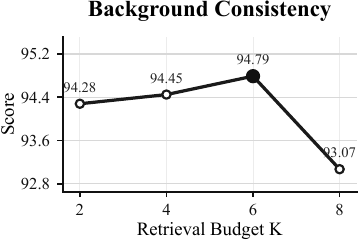} \\[-0.05em]
\includegraphics[width=0.46\linewidth]{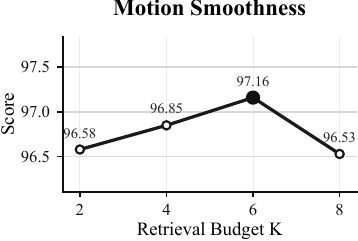} &
\includegraphics[width=0.46\linewidth]{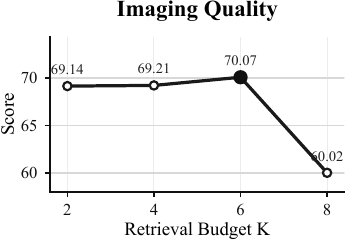}
\end{tabular}
\end{minipage}
\caption{
Ablation study and auxiliary VLM evaluation.
Left: embedding-space ablation and auxiliary VLM scores on 30s generations.
Right: retrieval-budget trends under the same total attention budget.
}
\label{tab:ablation_embedding}
\label{tab:ablation_budget}
\label{tab:vlm_eval}
\end{table*}

\paragraph{Embedding space.}
Table~\ref{tab:ablation_embedding} first tests the retrieval embedding space.
Learned embeddings outperform random retrieval, average-pooled descriptors, and reconstruction-only embeddings.
Adding $\mathcal{L}_{\mathrm{SeqDelta}}$ improves background consistency and motion smoothness over AE-only, while the full objective gives the strongest overall results across the retained metrics.
These trends match the design in Section~\ref{subsec:temporal_constraints}: retrieval benefits from a learned space that is both content-preserving and discriminative for non-local context.

\paragraph{Retrieval budget.}
With the embedding fixed, the retrieval-budget plots in Table~\ref{tab:ablation_embedding} show that $K=6$ gives the strongest consistency and imaging quality under the same total attention budget.
This setting allocates enough slots to retrieved non-local context while preserving local context for smooth continuation.
It therefore provides a practical balance between long-range reference and video continuity.
Appendix~\ref{app:ablation_examples} provides additional examples.

\section{Conclusion}
\label{sec:conclusion}

We presented LongLive-RAG, a simple way for an AR video generator to look back at useful parts of the video it has already generated.
Instead of relying only on the recent window, LongLive-RAG searches the generated history and brings back relevant context for the next block.
We train the retrieval embeddings with reconstruction, Window Temporal Delta Loss, and smoothing so that the search keys preserve visual content, avoid redundant nearby matches, and remain stable over time.
Experiments across multiple AR backbones and video lengths show that this improves long-video quality with little retrieval overhead.
Limitations are discussed in Appendix~\ref{sec:limitations}.

\bibliographystyle{plainnat}
\bibliography{main}

@article{videogpt,
  title={Videogpt: Video generation using vq-vae and transformers},
  author={Yan, Wilson and Zhang, Yunzhi and Abbeel, Pieter and Srinivas, Aravind},
  journal={arXiv preprint arXiv:2104.10157},
  year={2021}
}

@inproceedings{magvit,
  title={Magvit: Masked generative video transformer},
  author={Yu, Lijun and Cheng, Yong and Sohn, Kihyuk and Lezama, Jos{\'e} and Zhang, Han and Chang, Huiwen and Hauptmann, Alexander G and Yang, Ming-Hsuan and Hao, Yuan and Essa, Irfan and others},
  booktitle={Proceedings of the IEEE/CVF Conference on Computer Vision and Pattern Recognition},
  pages={10459--10469},
  year={2023}
}

@inproceedings{
magvitv2,
title={Language Model Beats Diffusion - Tokenizer is key to visual generation},
author={Lijun Yu and Jose Lezama and Nitesh Bharadwaj Gundavarapu and Luca Versari and Kihyuk Sohn and David Minnen and Yong Cheng and Agrim Gupta and Xiuye Gu and Alexander G Hauptmann and Boqing Gong and Ming-Hsuan Yang and Irfan Essa and David A Ross and Lu Jiang},
booktitle={The Twelfth International Conference on Learning Representations},
year={2024},
url={https://openreview.net/forum?id=gzqrANCF4g}
}

@article{videopoet,
  title={Videopoet: A large language model for zero-shot video generation},
  author={Kondratyuk, Dan and Yu, Lijun and Gu, Xiuye and Lezama, Jos{\'e} and Huang, Jonathan and Schindler, Grant and Hornung, Rachel and Birodkar, Vighnesh and Yan, Jimmy and Chiu, Ming-Chang and others},
  journal={arXiv preprint arXiv:2312.14125},
  year={2023}
}

@inproceedings{dino,
  title={Emerging Properties in Self-Supervised Vision Transformers},
  author={Caron, Mathilde and Touvron, Hugo and Misra, Ishan and Jegou, Herve and Mairal, Julien and Bojanowski, Piotr and Joulin, Armand},
  booktitle={Proceedings of the IEEE/CVF International Conference on Computer Vision},
  pages={9650--9660},
  year={2021}
}

@inproceedings{
nova,
title={Autoregressive Video Generation without Vector Quantization},
author={Haoge Deng and Ting Pan and Haiwen Diao and Zhengxiong Luo and Yufeng Cui and Huchuan Lu and Shiguang Shan and Yonggang Qi and Xinlong Wang},
booktitle={The Thirteenth International Conference on Learning Representations},
year={2025},
url={https://openreview.net/forum?id=JE9tCwe3lp}
}

@inproceedings{
pyramidalflow,
title={Pyramidal Flow Matching for Efficient Video Generative Modeling},
author={Yang Jin and Zhicheng Sun and Ningyuan Li and Kun Xu and Kun Xu and Hao Jiang and Nan Zhuang and Quzhe Huang and Yang Song and Yadong MU and Zhouchen Lin},
booktitle={The Thirteenth International Conference on Learning Representations},
year={2025},
url={https://openreview.net/forum?id=66NzcRQuOq}
}

@article{magi1,
  title={Magi-1: Autoregressive video generation at scale},
  author={Teng, Hansi and Jia, Hongyu and Sun, Lei and Li, Lingzhi and Li, Maolin and Tang, Mingqiu and Han, Shuai and Zhang, Tianning and Zhang, WQ and Luo, Weifeng and others},
  journal={arXiv preprint arXiv:2505.13211},
  year={2025}
}

@article{skyreels,
  title={Skyreels-v2: Infinite-length film generative model},
  author={Chen, Guibin and Lin, Dixuan and Yang, Jiangping and Lin, Chunze and Zhu, Junchen and Fan, Mingyuan and Zhang, Hao and Chen, Sheng and Chen, Zheng and Ma, Chengcheng and others},
  journal={arXiv preprint arXiv:2504.13074},
  year={2025}
}

@article{streamdit,
  title={Streamdit: Real-time streaming text-to-video generation},
  author={Kodaira, Akio and Hou, Tingbo and Hou, Ji and Georgopoulos, Markos and Juefei-Xu, Felix and Tomizuka, Masayoshi and Zhao, Yue},
  journal={arXiv preprint arXiv:2507.03745},
  year={2025}
}

@inproceedings{causvid,
  title={From slow bidirectional to fast autoregressive video diffusion models},
  author={Yin, Tianwei and Zhang, Qiang and Zhang, Richard and Freeman, William T and Durand, Fredo and Shechtman, Eli and Huang, Xun},
  booktitle={Proceedings of the IEEE/CVF Conference on Computer Vision and Pattern Recognition},
  pages={22963--22974},
  year={2025}
}

@inproceedings{
selfforcing,
title={Self Forcing: Bridging the Train-Test Gap in Autoregressive Video Diffusion},
author={Xun Huang and Zhengqi Li and Guande He and Mingyuan Zhou and Eli Shechtman},
booktitle={The Thirty-ninth Annual Conference on Neural Information Processing Systems},
year={2025},
url={https://openreview.net/forum?id=mSiN7i0BYH}
}

@article{selfforcingpp,
  title={Self-forcing++: Towards minute-scale high-quality video generation},
  author={Cui, Justin and Wu, Jie and Li, Ming and Yang, Tao and Li, Xiaojie and Wang, Rui and Bai, Andrew and Ban, Yuanhao and Hsieh, Cho-Jui},
  journal={arXiv preprint arXiv:2510.02283},
  year={2025}
}

@article{rollingforcing,
  title={Rolling forcing: Autoregressive long video diffusion in real time},
  author={Liu, Kunhao and Hu, Wenbo and Xu, Jiale and Shan, Ying and Lu, Shijian},
  journal={arXiv preprint arXiv:2509.25161},
  year={2025}
}

@article{causalforcing,
  title={Causal Forcing: Autoregressive Diffusion Distillation Done Right for High-Quality Real-Time Interactive Video Generation},
  author={Zhu, Hongzhou and Zhao, Min and He, Guande and Su, Hang and Li, Chongxuan and Zhu, Jun},
  journal={arXiv preprint arXiv:2602.02214},
  year={2026}
}

@inproceedings{
longlive,
title={LongLive: Real-time Interactive Long Video Generation},
author={Shuai Yang and Wei Huang and Ruihang Chu and Yicheng Xiao and Yuyang Zhao and Xianbang Wang and Muyang Li and Enze Xie and Ying-Cong Chen and Yao Lu and Song Han and Yukang Chen},
booktitle={The Fourteenth International Conference on Learning Representations},
year={2026},
url={https://openreview.net/forum?id=nCAODkpsPJ}
}

@inproceedings{
streamingllm,
title={Efficient Streaming Language Models with Attention Sinks},
author={Guangxuan Xiao and Yuandong Tian and Beidi Chen and Song Han and Mike Lewis},
booktitle={The Twelfth International Conference on Learning Representations},
year={2024},
url={https://openreview.net/forum?id=NG7sS51zVF}
}

@article{deepforcing,
  title={Deep forcing: Training-free long video generation with deep sink and participative compression},
  author={Yi, Jung and Jang, Wooseok and Cho, Paul Hyunbin and Nam, Jisu and Yoon, Heeji and Kim, Seungryong},
  journal={arXiv preprint arXiv:2512.05081},
  year={2025}
}

@article{contextforcing,
  title={Context Forcing: Consistent Autoregressive Video Generation with Long Context},
  author={Chen, Shuo and Wei, Cong and Sun, Sun and Nie, Ping and Zhou, Kai and Zhang, Ge and Yang, Ming-Hsuan and Chen, Wenhu},
  journal={arXiv preprint arXiv:2602.06028},
  year={2026}
}

@inproceedings{streamingt2v,
  title={Streamingt2v: Consistent, dynamic, and extendable long video generation from text},
  author={Henschel, Roberto and Khachatryan, Levon and Poghosyan, Hayk and Hayrapetyan, Daniil and Tadevosyan, Vahram and Wang, Zhangyang and Navasardyan, Shant and Shi, Humphrey},
  booktitle={Proceedings of the Computer Vision and Pattern Recognition Conference},
  pages={2568--2577},
  year={2025}
}

@article{rope,
  title={Roformer: Enhanced transformer with rotary position embedding},
  author={Su, Jianlin and Ahmed, Murtadha and Lu, Yu and Pan, Shengfeng and Bo, Wen and Liu, Yunfeng},
  journal={Neurocomputing},
  volume={568},
  pages={127063},
  year={2024},
  publisher={Elsevier}
}

@article{infinityrope,
  title={Infinity-rope: Action-controllable infinite video generation emerges from autoregressive self-rollout},
  author={Yesiltepe, Hidir and Meral, Tuna Han Salih and Akan, Adil Kaan and Oktay, Kaan and Yanardag, Pinar},
  journal={arXiv preprint arXiv:2511.20649},
  year={2025}
}

@article{lol,
  title={LoL: Longer than Longer, Scaling Video Generation to Hour},
  author={Cui, Justin and Wu, Jie and Li, Ming and Yang, Tao and Li, Xiaojie and Wang, Rui and Bai, Andrew and Ban, Yuanhao and Hsieh, Cho-Jui},
  journal={arXiv preprint arXiv:2601.16914},
  year={2026}
}

@inproceedings{
riflex,
title={{RIFLE}x: A Free Lunch for Length Extrapolation in Video Diffusion Transformers},
author={Min Zhao and Guande He and Yixiao Chen and Hongzhou Zhu and Chongxuan Li and Jun Zhu},
booktitle={Forty-second International Conference on Machine Learning},
year={2025},
url={https://openreview.net/forum?id=v3B79m7t8Z}
}

@article{wan,
  title={Wan: Open and advanced large-scale video generative models},
  author={Wan, Team and Wang, Ang and Ai, Baole and Wen, Bin and Mao, Chaojie and Xie, Chen-Wei and Chen, Di and Yu, Feiwu and Zhao, Haiming and Yang, Jianxiao and others},
  journal={arXiv preprint arXiv:2503.20314},
  year={2025}
}

@inproceedings{vbench,
  title={Vbench: Comprehensive benchmark suite for video generative models},
  author={Huang, Ziqi and He, Yinan and Yu, Jiashuo and Zhang, Fan and Si, Chenyang and Jiang, Yuming and Zhang, Yuanhan and Wu, Tianxing and Jin, Qingyang and Chanpaisit, Nattapol and others},
  booktitle={Proceedings of the IEEE/CVF Conference on Computer Vision and Pattern Recognition},
  pages={21807--21818},
  year={2024}
}

@article{moviegen,
  title={Movie gen: A cast of media foundation models},
  author={Polyak, Adam and Zohar, Amit and Brown, Andrew and Tjandra, Andros and Sinha, Animesh and Lee, Ann and Vyas, Apoorv and Shi, Bowen and Ma, Chih-Yao and Chuang, Ching-Yao and others},
  journal={arXiv preprint arXiv:2410.13720},
  year={2024}
}

@article{qwen25,
  title={Qwen2.5 Technical Report},
  author={{Qwen Team}},
  journal={arXiv preprint arXiv:2412.15115},
  year={2024}
}

@article{diffusionforcing,
  title={Diffusion forcing: Next-token prediction meets full-sequence diffusion},
  author={Chen, Boyuan and Mart{\'\i} Mons{\'o}, Diego and Du, Yilun and Simchowitz, Max and Tedrake, Russ and Sitzmann, Vincent},
  journal={Advances in Neural Information Processing Systems},
  volume={37},
  pages={24081--24125},
  year={2024}
}

@article{pi,
  title={Extending context window of large language models via positional interpolation},
  author={Chen, Shouyuan and Wong, Sherman and Chen, Liangjian and Tian, Yuandong},
  journal={arXiv preprint arXiv:2306.15595},
  year={2023}
}

@article{yarn,
  title={Yarn: Efficient context window extension of large language models},
  author={Peng, Bowen and Quesnelle, Jeffrey and Fan, Honglu and Shippole, Enrico},
  journal={arXiv preprint arXiv:2309.00071},
  year={2023}
}

@misc{ntk,
  author    = {bloc97},
  title     = {{NTK}-Aware Scaled {RoPE} allows {LLaMA} models to have extended (8k+) context size without any fine-tuning and minimal perplexity degradation},
  year      = {2023},
  howpublished = {Reddit post},
  url       = {https://www.reddit.com/r/LocalLLaMA/comments/14lz7j5/}
}

@article{yang2026stableworld,
  title={StableWorld: Towards Stable and Consistent Long Interactive Video Generation},
  author={Yang, Ying and Lv, Zhengyao and Pan, Tianlin and Wang, Haofan and Yang, Binxin and Yin, Hubery and Li, Chen and Liu, Ziwei and Si, Chenyang},
  journal={arXiv preprint arXiv:2601.15281},
  year={2026}
}

@article{sun2025worldplay,
  title={Worldplay: Towards long-term geometric consistency for real-time interactive world modeling},
  author={Sun, Wenqiang and Zhang, Haiyu and Wang, Haoyuan and Wu, Junta and Wang, Zehan and Wang, Zhenwei and Wang, Yunhong and Zhang, Jun and Wang, Tengfei and Guo, Chunchao},
  journal={arXiv preprint arXiv:2512.14614},
  year={2025}
}

@article{brooks2024video,
  title={Video generation models as world simulators},
  author={Brooks, Tim and Peebles, Bill and Holmes, Connor and DePue, Will and Guo, Yufei and Jing, Leo and Schnurr, David and Taylor, Joe and Luhman, Troy and Luhman, Eric and others},
  journal={OpenAI Blog},
  volume={1},
  number={8},
  pages={1},
  year={2024}
}

@article{hong2025relic,
  title={Relic: Interactive video world model with long-horizon memory},
  author={Hong, Yicong and Mei, Yiqun and Ge, Chongjian and Xu, Yiran and Zhou, Yang and Bi, Sai and Hold-Geoffroy, Yannick and Roberts, Mike and Fisher, Matthew and Shechtman, Eli and others},
  journal={arXiv preprint arXiv:2512.04040},
  year={2025}
}

@article{feng2024matrix,
  title={The matrix: Infinite-horizon world generation with real-time moving control},
  author={Feng, Ruili and Zhang, Han and Yang, Zhantao and Xiao, Jie and Shu, Zhilei and Liu, Zhiheng and Zheng, Andy and Huang, Yukun and Liu, Yu and Zhang, Hongyang},
  journal={arXiv preprint arXiv:2412.03568},
  year={2024}
}

@article{ren2025cosmos,
  title={Cosmos-drive-dreams: Scalable synthetic driving data generation with world foundation models},
  author={Ren, Xuanchi and Lu, Yifan and Cao, Tianshi and Gao, Ruiyuan and Huang, Shengyu and Sabour, Amirmojtaba and Shen, Tianchang and Pfaff, Tobias and Wu, Jay Zhangjie and Chen, Runjian and others},
  journal={arXiv preprint arXiv:2506.09042},
  year={2025}
}

@article{hu2023gaia,
  title={Gaia-1: A generative world model for autonomous driving},
  author={Hu, Anthony and Russell, Lloyd and Yeo, Hudson and Murez, Zak and Fedoseev, George and Kendall, Alex and Shotton, Jamie and Corrado, Gianluca},
  journal={arXiv preprint arXiv:2309.17080},
  year={2023}
}

@inproceedings{elmoghany2025survey,
  title={A survey on long-video storytelling generation: architectures, consistency, and cinematic quality},
  author={Elmoghany, Mohamed and Rossi, Ryan and Yoon, Seunghyun and Mukherjee, Subhojyoti and Bakr, Eslam Mohamed and Mathur, Puneet and Wu, Gang and Lai, Viet Dac and Lipka, Nedim and Zhang, Ruiyi and others},
  booktitle={Proceedings of the IEEE/CVF International Conference on Computer Vision},
  pages={7023--7035},
  year={2025}
}

@article{liu2024sora,
  title={Sora: A review on background, technology, limitations, and opportunities of large vision models},
  author={Liu, Yixin and Zhang, Kai and Li, Yuan and Yan, Zhiling and Gao, Chujie and Chen, Ruoxi and Yuan, Zhengqing and Huang, Yue and Sun, Hanchi and Gao, Jianfeng and others},
  journal={arXiv preprint arXiv:2402.17177},
  year={2024}
}

@article{kong2024hunyuanvideo,
  title={Hunyuanvideo: A systematic framework for large video generative models},
  author={Kong, Weijie and Tian, Qi and Zhang, Zijian and Min, Rox and Dai, Zuozhuo and Zhou, Jin and Xiong, Jiangfeng and Li, Xin and Wu, Bo and Zhang, Jianwei and others},
  journal={arXiv preprint arXiv:2412.03603},
  year={2024}
}

@article{yang2024cogvideox,
  title={Cogvideox: Text-to-video diffusion models with an expert transformer},
  author={Yang, Zhuoyi and Teng, Jiayan and Zheng, Wendi and Ding, Ming and Huang, Shiyu and Xu, Jiazheng and Yang, Yuanming and Hong, Wenyi and Zhang, Xiaohan and Feng, Guanyu and others},
  journal={arXiv preprint arXiv:2408.06072},
  year={2024}
}

@inproceedings{dmd,
  title={One-step diffusion with distribution matching distillation},
  author={Yin, Tianwei and Gharbi, Micha{\"e}l and Zhang, Richard and Shechtman, Eli and Durand, Fredo and Freeman, William T and Park, Taesung},
  booktitle={Proceedings of the IEEE/CVF conference on computer vision and pattern recognition},
  pages={6613--6623},
  year={2024}
}

@article{zhang2023h2o,
  title={H2o: Heavy-hitter oracle for efficient generative inference of large language models},
  author={Zhang, Zhenyu and Sheng, Ying and Zhou, Tianyi and Chen, Tianlong and Zheng, Lianmin and Cai, Ruisi and Song, Zhao and Tian, Yuandong and R{\'e}, Christopher and Barrett, Clark and others},
  journal={Advances in Neural Information Processing Systems},
  volume={36},
  pages={34661--34710},
  year={2023}
}

@article{li2024snapkv,
  title={Snapkv: Llm knows what you are looking for before generation},
  author={Li, Yuhong and Huang, Yingbing and Yang, Bowen and Venkitesh, Bharat and Locatelli, Acyr and Ye, Hanchen and Cai, Tianle and Lewis, Patrick and Chen, Deming},
  journal={Advances in Neural Information Processing Systems},
  volume={37},
  pages={22947--22970},
  year={2024}
}

@article{wan2024d2o,
  title={D2o: Dynamic discriminative operations for efficient generative inference of large language models},
  author={Wan, Zhongwei and Wu, Xinjian and Zhang, Yu and Xin, Yi and Tao, Chaofan and Zhu, Zhihong and Wang, Xin and Luo, Siqi and Xiong, Jing and Zhang, Mi},
  journal={arXiv preprint arXiv:2406.13035},
  volume={2},
  year={2024}
}

@article{ghadia2025dialogue,
  title={Dialogue without limits: Constant-sized KV caches for extended responses in LLMs},
  author={Ghadia, Ravi and Kumar, Avinash and Jain, Gaurav and Nair, Prashant and Das, Poulami},
  journal={arXiv preprint arXiv:2503.00979},
  year={2025}
}

@inproceedings{li2025vmem,
  title={Vmem: Consistent interactive video scene generation with surfel-indexed view memory},
  author={Li, Runjia and Torr, Philip and Vedaldi, Andrea and Jakab, Tomas},
  booktitle={Proceedings of the IEEE/CVF International Conference on Computer Vision},
  pages={25690--25699},
  year={2025}
}

@article{xiao2025worldmem,
  title={Worldmem: Long-term consistent world simulation with memory},
  author={Xiao, Zeqi and Lan, Yushi and Zhou, Yifan and Ouyang, Wenqi and Yang, Shuai and Zeng, Yanhong and Pan, Xingang},
  journal={arXiv preprint arXiv:2504.12369},
  year={2025}
}

@inproceedings{yu2025context,
  title={Context as memory: Scene-consistent interactive long video generation with memory retrieval},
  author={Yu, Jiwen and Bai, Jianhong and Qin, Yiran and Liu, Quande and Wang, Xintao and Wan, Pengfei and Zhang, Di and Liu, Xihui},
  booktitle={Proceedings of the SIGGRAPH Asia 2025 Conference Papers},
  pages={1--11},
  year={2025}
}

@article{yu2025videossm,
  title={Videossm: Autoregressive long video generation with hybrid state-space memory},
  author={Yu, Yifei and Wu, Xiaoshan and Hu, Xinting and Hu, Tao and Sun, Yangtian and Lyu, Xiaoyang and Wang, Bo and Ma, Lin and Ma, Yuewen and Wang, Zhongrui and others},
  journal={arXiv preprint arXiv:2512.04519},
  year={2025}
}

@article{zhang2025pretraining,
  title={Pretraining Frame Preservation in Autoregressive Video Memory Compression},
  author={Zhang, Lvmin and Cai, Shengqu and Li, Muyang and Zeng, Chong and Lu, Beijia and Rao, Anyi and Han, Song and Wetzstein, Gordon and Agrawala, Maneesh},
  journal={arXiv preprint arXiv:2512.23851},
  year={2025}
}

@article{zhang2025packing,
  title={Packing input frame context in next-frame prediction models for video generation},
  author={Zhang, Lvmin and Agrawala, Maneesh},
  journal={arXiv e-prints},
  pages={arXiv--2504},
  year={2025}
}

@article{wu2026infinite,
  title={Infinite-World: Scaling Interactive World Models to 1000-Frame Horizons via Pose-Free Hierarchical Memory},
  author={Wu, Ruiqi and He, Xuanhua and Cheng, Meng and Yang, Tianyu and Zhang, Yong and Kang, Zhuoliang and Cai, Xunliang and Wei, Xiaoming and Guo, Chunle and Li, Chongyi and others},
  journal={arXiv preprint arXiv:2602.02393},
  year={2026}
}

@article{gu2025long,
  title={Long-context autoregressive video modeling with next-frame prediction},
  author={Gu, Yuchao and Mao, Weijia and Shou, Mike Zheng},
  journal={arXiv preprint arXiv:2503.19325},
  year={2025}
}

@inproceedings{zhai2025stargen,
  title={Stargen: A spatiotemporal autoregression framework with video diffusion model for scalable and controllable scene generation},
  author={Zhai, Shangjin and Ye, Zhichao and Liu, Jialin and Xie, Weijian and Hu, Jiaqi and Peng, Zhen and Xue, Hua and Chen, Danpeng and Wang, Xiaomeng and Yang, Lei and others},
  booktitle={Proceedings of the Computer Vision and Pattern Recognition Conference},
  pages={26822--26833},
  year={2025}
}

@article{wang2024model,
  title={Model tells you where to merge: Adaptive kv cache merging for llms on long-context tasks},
  author={Wang, Zheng and Jin, Boxiao and Yu, Zhongzhi and Zhang, Minjia},
  journal={arXiv preprint arXiv:2407.08454},
  year={2024}
}

@inproceedings{zhang2024cam,
  title={Cam: Cache merging for memory-efficient llms inference},
  author={Zhang, Yuxin and Du, Yuxuan and Luo, Gen and Zhong, Yunshan and Zhang, Zhenyu and Liu, Shiwei and Ji, Rongrong},
  booktitle={Forty-first international conference on machine learning},
  year={2024}
}

@misc{gemini31pro,
  title={Gemini 3.1 Pro Model Card},
  author={{Google DeepMind}},
  year={2026},
  url={https://deepmind.google/models/model-cards/gemini-3-1-pro/}
}

@article{kim2026memrope,
  title={MemRoPE: Training-Free Infinite Video Generation via Evolving Memory Tokens},
  author={Kim, Youngrae and Hu, Qixin and Kuo, C-C Jay and Beerel, Peter A},
  journal={arXiv preprint arXiv:2603.12513},
  year={2026}
}

@article{li2026rolling,
  title={Rolling Sink: Bridging Limited-Horizon Training and Open-Ended Testing in Autoregressive Video Diffusion},
  author={Li, Haodong and Liu, Shaoteng and Lin, Zhe and Chandraker, Manmohan},
  journal={arXiv preprint arXiv:2602.07775},
  year={2026}
}

@article{lu2025reward,
  title={Reward forcing: Efficient streaming video generation with rewarded distribution matching distillation},
  author={Lu, Yunhong and Zeng, Yanhong and Li, Haobo and Ouyang, Hao and Wang, Qiuyu and Cheng, Ka Leong and Zhu, Jiapeng and Cao, Hengyuan and Zhang, Zhipeng and Zhu, Xing and others},
  journal={arXiv preprint arXiv:2512.04678},
  year={2025}
}

@inproceedings{lewis2020rag,
  title={Retrieval-Augmented Generation for Knowledge-Intensive NLP Tasks},
  author={Lewis, Patrick and Perez, Ethan and Piktus, Aleksandra and Petroni, Fabio and Karpukhin, Vladimir and Goyal, Naman and Kuttler, Heinrich and Lewis, Mike and Yih, Wen-tau and Rockt{\"a}schel, Tim and Riedel, Sebastian and Kiela, Douwe},
  booktitle={Advances in Neural Information Processing Systems},
  volume={33},
  pages={9459--9474},
  year={2020}
}

@inproceedings{guu2020realm,
  title={{REALM}: Retrieval-Augmented Language Model Pre-Training},
  author={Guu, Kelvin and Lee, Kenton and Tung, Zora and Pasupat, Panupong and Chang, Ming-Wei},
  booktitle={Proceedings of the 37th International Conference on Machine Learning},
  pages={3929--3938},
  year={2020}
}

@inproceedings{borgeaud2022retro,
  title={Improving Language Models by Retrieving from Trillions of Tokens},
  author={Borgeaud, Sebastian and Mensch, Arthur and Hoffmann, Jordan and Cai, Trevor and Rutherford, Eliza and Millican, Katie and van den Driessche, George and Lespiau, Jean-Baptiste and Damoc, Bogdan and Clark, Aidan and others},
  booktitle={Proceedings of the 39th International Conference on Machine Learning},
  pages={2206--2240},
  year={2022}
}

%%%%%%%%%%%%%%%%%%%%%%%%%%%%%%%%%%%%%%%%%%%%%%%%%%%%%%%%%%%%

\newpage
\appendix

\section{Limitations}
\label{sec:limitations}

LongLive-RAG builds on a frozen base checkpoint.
It improves how the model selects and reuses generated history, but it does not change the generator itself.
As a result, the final video quality is still bounded by the capability of the base AR model.

\section{Broader Impacts}
\label{sec:broader_impacts}

LongLive-RAG is a research method for improving long-horizon consistency in AR text-to-video generation.
It may support positive applications that require stable long video synthesis, such as creative tools, simulation, and interactive content generation.
At the same time, improvements in long-video quality can inherit the misuse risks of the underlying video generator, including deceptive or misleading generated media.
LongLive-RAG does not introduce a new base generator or a new data source; it changes how a frozen AR generator selects and reuses its generated history.
Responsible use should therefore follow the license, usage terms, and safety practices of the underlying video generation models.

\section{Additional Related Work Discussion}
\label{app:related_work}

\paragraph{Video generation and AR rollout.}
Modern video generators include token-based models, diffusion transformers, and large-scale text-to-video systems \citep{videogpt,magvit,magvitv2,videopoet,nova,pyramidalflow,wan,moviegen,kong2024hunyuanvideo,yang2024cogvideox,magi1,skyreels}.
Many of these models denoise a fixed clip jointly, which ties computation to clip length.
AR video generation instead emits frames or latent blocks causally, enabling streaming, interactive prompting, and variable-length output \citep{diffusionforcing,causvid,selfforcing,rollingforcing,longlive,streamdit,streamingt2v,gu2025long,zhai2025stargen}.
LongLive-RAG is designed for this AR setting, where the generated trajectory itself becomes the source of future context.

\paragraph{Causal adaptation and self-generated context.}
Several methods convert or adapt pretrained video diffusion models into causal generators through distillation, causal teacher--student training, or self-generated context exposure \citep{dmd,causvid,selfforcing,selfforcingpp,causalforcing,contextforcing,rollingforcing,longlive,lol}.
These techniques reduce exposure bias and improve each local denoising step under causal rollout.
LongLive-RAG addresses a complementary axis: given a causal generator and its self-generated trajectory, how should non-local context be selected at inference time?
It leaves the denoiser unchanged and modifies only the context assembly policy.

\paragraph{Context windows, anchors, and positional extrapolation.}
The simplest scalable policy is a sliding window, which bounds memory by keeping only recent context \citep{causvid,selfforcing,rollingforcing,longlive}.
Attention sinks and anchor-based variants preserve early tokens or frames to improve stability, but the chosen anchors are not necessarily the content needed by a later query \citep{streamingllm,longlive,li2026rolling,lu2025reward}.
A related line extends positional encodings so that models can operate beyond their trained length, including RoPE-based positional extrapolation and video-specific variants \citep{rope,pi,yarn,ntk,riflex,infinityrope}.
These methods increase the feasible context horizon or stabilize attention, but their selection rule is still primarily position-based.
LongLive-RAG instead uses content-addressed retrieval to decide which historical entries are exposed to attention.

\paragraph{Compressed-history tokens and recurrent history.}
Methods based on compressed-history tokens and cache management summarize, select, or evict long histories through substitute tokens, recurrent states, compact sequence representations, or KV-cache policies \citep{zhang2023h2o,li2024snapkv,wan2024d2o,ghadia2025dialogue,deepforcing,yu2025videossm,kim2026memrope,zhang2025pretraining,zhang2025packing,wang2024model,zhang2024cam}.
Such designs are attractive because they decouple memory cost from the full history length.
Their limitation for generation is that the model often attends to the summary itself rather than to the original context produced by the generator.
If the summary drops a rare object, identity cue, or background detail, a later denoising step cannot recover it directly.
LongLive-RAG also uses compact representations, but only as search keys; after retrieval, the generator receives matched context entries in the backbone's native cache format.
This separates the compression needed for indexing from the information used for denoising.

\paragraph{Retrieval-augmented generation.}
Retrieval-augmented generation retrieves external evidence or memory and conditions a generator on the retrieved content \citep{lewis2020rag,guu2020realm,borgeaud2022retro}.
These methods are typically used with a fixed text corpus or database to improve factual grounding or long-context access.
LongLive-RAG uses the same high-level retrieval principle, but the retrieval source is different: it searches self-generated video latents produced during the current AR rollout.
The retrieved items are also inserted as native generator context rather than as external text evidence.
This makes retrieval a context-selection mechanism for long video generation.

\paragraph{Retrieval memory in video generation.}
Retrieval memory has been explored in world models and embodied video settings where explicit structure is available.
Memory can be organized by camera pose, 3D geometry, scene coordinates, or field-of-view overlap, making it possible to identify relevant observations in a physically grounded way \citep{hu2023gaia,feng2024matrix,li2025vmem,yu2025context,xiao2025worldmem,sun2025worldplay,yang2026stableworld,wu2026infinite}.
Open-ended text-to-video generation does not provide such explicit retrieval cues by default: the camera can move freely, objects can reappear after occlusion, and no pose graph indicates which historical state should be reused.
LongLive-RAG therefore retrieves from the generator's own latent trajectory and learns the search geometry from generated content.

\section{Why Retrieval in Latent Space?}
\label{app:latent_vs_pixel_retrieval}

Most contemporary video diffusion systems denoise in a VAE latent space and decode only after the latent rollout is complete \citep{wan,moviegen,kong2024hunyuanvideo,yang2024cogvideox}.
A raw-pixel retrieval pipeline would break this execution pattern.
After every completed AR block, the system would need to immediately run VAE decoding before search, then either store decoded frames or extract pixel-space features for all historical candidates.
This inserts an extra decode-and-transfer path into each generation step, increasing memory traffic and communication overhead relative to the standard latent-first pipeline.

\begin{figure}[h]
    \centering
    \includegraphics[width=0.78\linewidth]{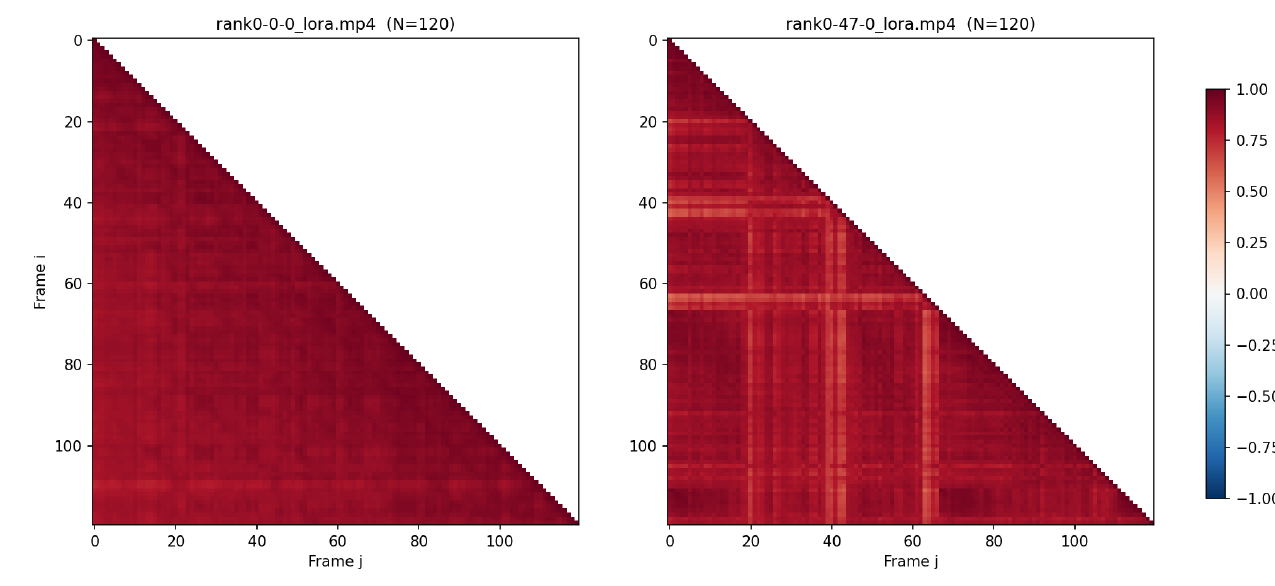}
    \caption{
    Raw-pixel retrieval analysis using DINO features~\citep{dino} from decoded frames.
    The similarity map shows that off-the-shelf image features tend to retrieve temporally local neighbors, making them less suitable for long generated videos where useful context may be far outside the recent window.
    }
    \label{fig:app_dinosim}
\end{figure}

Latent retrieval also gives a simpler learning problem.
The generator's latents already carry visual and semantic information in a representation aligned with the denoising model and its attention context.
Training a lightweight encoder on these latents can therefore focus on making the search space discriminative for recurrence.
In contrast, a pixel-space retriever must learn from decoded frames, which are higher-dimensional and less directly tied to the generator's internal state.
Using an off-the-shelf image feature space is not a clean substitute: such features can be semantically broad but insufficiently discriminative for long generated videos, causing the nearest neighbors to remain overly local in time, similar to the collapse observed with reconstruction-only latent compression.
Figure~\ref{fig:app_dinosim} illustrates this behavior with DINO features~\citep{dino} computed from decoded frames: the similarity structure is dominated by local neighborhoods rather than clean long-range matches.

Finally, the mapping between latent states and decoded pixels is not one-to-one for retrieval.
The same or similar decoded appearance can correspond to different latent states that carry different denoising histories, context associations, or future trajectory implications.
A pixel-space match can therefore be ambiguous when mapped back to the latent/context objects needed by the AR generator.
LongLive-RAG avoids this mismatch by indexing the objects the generator actually produces and reuses: self-generated latents and their original context.

\section{Experiment Details}
\label{app:experimental_details}

This appendix section records the experimental details used for LongLive-RAG.

\paragraph{Training data construction.}
We use the prompt pool from Self-Forcing \citep{selfforcing}.
We randomly sample $10\%$ of the prompts with seed 0.
For each selected prompt, we generate 90-latent LongLive rollouts with a frozen generator \citep{longlive}.
We use LongLive because it provides long AR rollout latents with history-dependent states in the target latent space.
This choice is used only for collecting training latents for the retrieval encoder; it is not used to filter evaluation prompts or select reported samples.
We save the clean denoised latent from each completed AR block and use these latents to train the autoencoder.
Generating the resulting 20,000+ training latents takes about 7 days on 6 NVIDIA RTX A6000 GPUs.
This data-construction stage is separate from the inference and evaluation runs reported in Section~\ref{subsec:exp_setup}.
Decoded videos are used only for sanity checking.

\paragraph{Why single base model?}
\textbf{Our choice is to train one retrieval encoder and use it unchanged across all evaluated backbones.}
A natural alternative is to train a separate retrieval encoder for each target AR backbone, or to train on rollouts collected from all base models.
We deliberately avoid this setting.
LongLive-RAG is designed as a general RAG framework; per-backbone retrieval would make the method tied to each new AR generator.
We therefore train one encoder in the shared WAN VAE latent space used by many recent AR video backbones, including Causal-Forcing, Self-Forcing, and LongLive \citep{causalforcing,selfforcing,longlive}.
Although these backbones may have different rollout distributions, their generated latents are decoded by the same WAN VAE and therefore share a common latent coordinate system.
The same frozen encoder is used for all evaluated backbones in Table~\ref{tab:main}, without per-backbone retraining or data-size tuning.

\paragraph{Latent autoencoder training.}
The base video generators are frozen, and only the latent autoencoder is trained.
The autoencoder is a convolutional encoder--decoder applied to each latent block independently.
The encoder uses stride-2 convolutions, GroupNorm, SiLU activations, residual blocks, global average pooling, and a linear projection to the retrieval embedding.
The decoder maps the embedding back to a bottleneck feature map and reconstructs the latent with nearest-neighbor upsampling, convolutions, GroupNorm, SiLU activations, and residual blocks.
The decoder has no output activation, and the output is cropped to the target latent size.
For retrieval, we use the encoder output with L2 normalization.

We train with AdamW.
The objective follows Section~\ref{subsec:temporal_constraints}: mean-squared reconstruction loss, Window Temporal Delta Loss, and trajectory-smoothing loss.
We split the latent dataset into training and validation sets, using 10\% for validation.
We track reconstruction, sequence-delta, smoothness, and total loss on both splits.
The best checkpoint is selected by the lowest validation total loss.
The retrieval autoencoder training job takes less than 100 NVIDIA RTX A6000 GPU hours after the training latents have been generated.

Table~\ref{tab:experimental_hparams} summarizes the main hyperparameters.
The values are read from the AE training config and the LongLive-RAG inference configs.

\begingroup
\scriptsize
\setlength{\tabcolsep}{6pt}
\renewcommand{\arraystretch}{0.98}

\begin{table}[!htbp]
\centering
\caption{Main hyperparameters used in LongLive-RAG.}
\label{tab:experimental_hparams}
\begin{tabular}{@{}p{0.28\linewidth}p{0.20\linewidth}p{0.42\linewidth}@{}}
\toprule
\textbf{parameter} & \textbf{value} & \textbf{description} \\
\midrule
Optimizer & AdamW & optimizer for training the latent autoencoder \\
Sequence length & 8 & continuous latent blocks per training chunk \\
Embedding dimension & 1024 & retrieval embedding dimension \\
Hidden dimensions & [64, 128, 256] & encoder and decoder channel dimensions \\
Learning rate & $3\times10^{-4}$ & learning rate for AdamW \\
Weight decay & $1\times10^{-4}$ & optimizer weight decay \\
Batch size & 128 & training batch size \\
Training epochs & 400 & total training epochs \\
$\lambda_{\mathrm{rec}}$ & 1 & reconstruction loss weight \\
$\lambda_{\Delta}$ & 1.0 & Window Temporal Delta Loss weight \\
Delta window $w$ & 3 & local window for temporal delta pairs \\
Delta margin $m$ & 0.85 & cosine-similarity margin \\
$\lambda_{\mathrm{smooth}}$ & 1 & trajectory-smoothing loss weight \\
Gradient clipping & 1.0 & maximum gradient norm \\
Mixed precision & enabled & mixed-precision training \\
Checkpoint selection & lowest val. total loss & validation total is the sum of tracked loss terms \\
Recency guard $R$ & 5 & number of recent blocks excluded from retrieval \\
\bottomrule
\end{tabular}
\end{table}
\endgroup

\paragraph{Evaluation protocol.}
We evaluate generated videos with the official VBench-Long scripts \citep{vbench}.
The reported inference and evaluation runs for Table~\ref{tab:main} and Table~\ref{tab:ablation_embedding} take about one week on 6 NVIDIA RTX A6000 GPUs, excluding the retrieval-encoder data-construction and autoencoder-training stages described above.
Runtime overhead is measured on one NVIDIA RTX A6000, with latent encoding and top-$K$ search timed separately.
We also conduct an auxiliary VLM evaluation on 30s generations.

\FloatBarrier

\section{Auxiliary VLM Evaluation}
\label{app:vlm_eval}

\paragraph{VLM evaluation.}
We use Gemini 3.1-Pro~\citep{gemini31pro} as the VLM judge to score each generated video.
Following Deep Forcing and Self-Forcing++~\citep{deepforcing,selfforcingpp}, we randomly sample 20 cases from the 30s generations for this auxiliary evaluation.
The VLM is given the generated video and returns an aggregate score following the prompt below.
Based on the original VLM scoring setup, we revise the system prompt by adding system instructions, a scoring example, and an explicit output format.
We average scores over the evaluated prompt set and report the mean in Table~\ref{tab:vlm_eval} as an auxiliary check.

\begin{tcolorbox}[
    enhanced,
    breakable,
    colback=white,
    colframe=gray!35,
    colbacktitle=gray!45,
    coltitle=white,
    title={VLM Evaluation Prompt},
    fonttitle=\large\bfseries,
    boxrule=1pt,
    arc=3pt,
    left=1.2em,
    right=1.2em,
    top=0.9em,
    bottom=0.9em,
    toptitle=0.5em,
    bottomtitle=0.5em
]
\small
You are an expert video quality assessor. Your task is to evaluate the ``Overall Exposure Quality'' of the provided video file.

Since exposure can change over time, you must watch the entire video and provide a single aggregate score based on the severity and duration of any exposure issues.

\medskip
\noindent\textbf{Scoring Scale}

\smallskip
\noindent\textbf{0: Catastrophic.} The video is dominated by pure white (blown out) or pure black (crushed) frames for a significant duration, rendering the content largely unreadable.

\noindent\textbf{1: Severe.} Large portions of the video suffer from severe over- or under-exposure, substantially impairing visibility and ruining the viewer's experience.

\noindent\textbf{2: Noticeable.} Persistent clipping in highlights or shadows throughout the video, or frequent noticeable shifts that cause a significant loss of detail.

\noindent\textbf{3: Moderate.} Exposure is generally acceptable, but contains brief periods of over/under-exposure, or constant but minor highlight/shadow clipping that preserves core visual information.

\noindent\textbf{4: Minor.} Exposure is mostly excellent. Small regions are occasionally too bright or dark, or there is a very brief, negligible fluctuation, but it does not disrupt visibility.

\noindent\textbf{5: Perfect.} Rock-solid, well-balanced lighting across the entire frame and duration. No distracting over-exposure, darkening, or unnatural shifts.

\medskip
\noindent\textbf{Evaluation Guidelines}

\smallskip
\noindent If the video has mixed quality (e.g., mostly a ``5'' but has a 2-second period of ``1''), penalize the overall score based on how distracting that bad period is (e.g., an overall score of 2 or 3).

\noindent Pay attention to both the spatial extent (how much of the screen is affected) and the temporal extent (how long the issue lasts).

\medskip
\noindent\textbf{Output Format}

\smallskip
\noindent First, provide a brief analysis summarizing the exposure across the video's duration (mentioning any specific moments of over/under-exposure if they exist). Then, output the final integer score wrapped in \texttt{\textless score\textgreater\textless/score\textgreater} tags.

\smallskip
\noindent\textbf{Example:}

\noindent Analysis: The first 5 seconds are perfectly exposed, but when the camera pans to the window, the sky becomes completely blown out for the remainder of the video, losing all highlight detail.

\noindent Score: \texttt{\textless score\textgreater3\textless/score\textgreater}
\end{tcolorbox}

\FloatBarrier

\section{Ablation Examples}
\label{app:ablation_examples}

We provide qualitative examples for the retrieval-budget and embedding-space ablations.

\begin{figure}[!htbp]
    \centering
    \includegraphics[width=\linewidth]{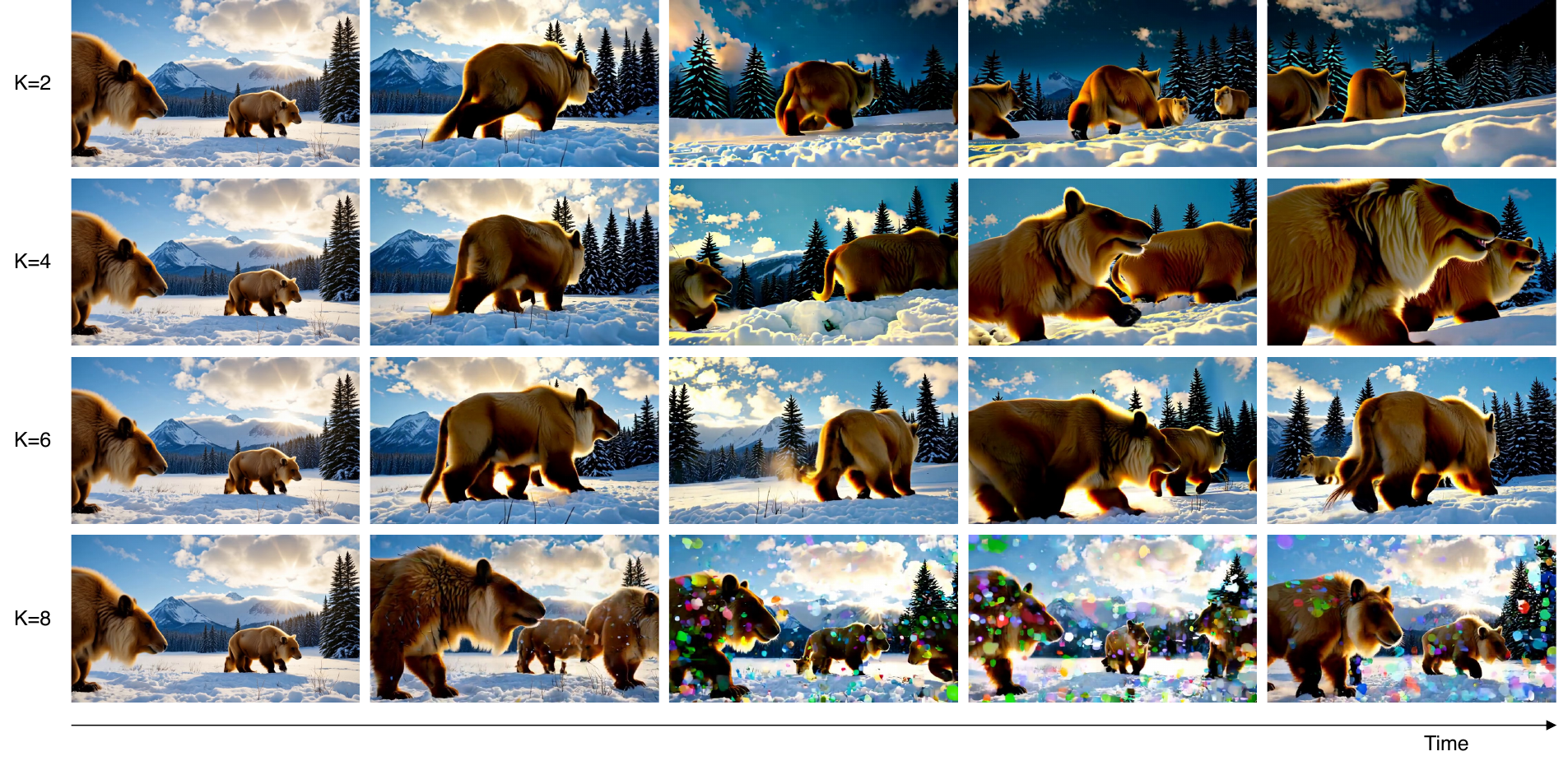}
    \caption{
    Retrieval-budget comparison.
    The balanced setting preserves both long-range references and video continuity.
    }
    \label{fig:app_ablationK}
\end{figure}

\begin{figure}[!htbp]
    \centering
    \includegraphics[width=\linewidth]{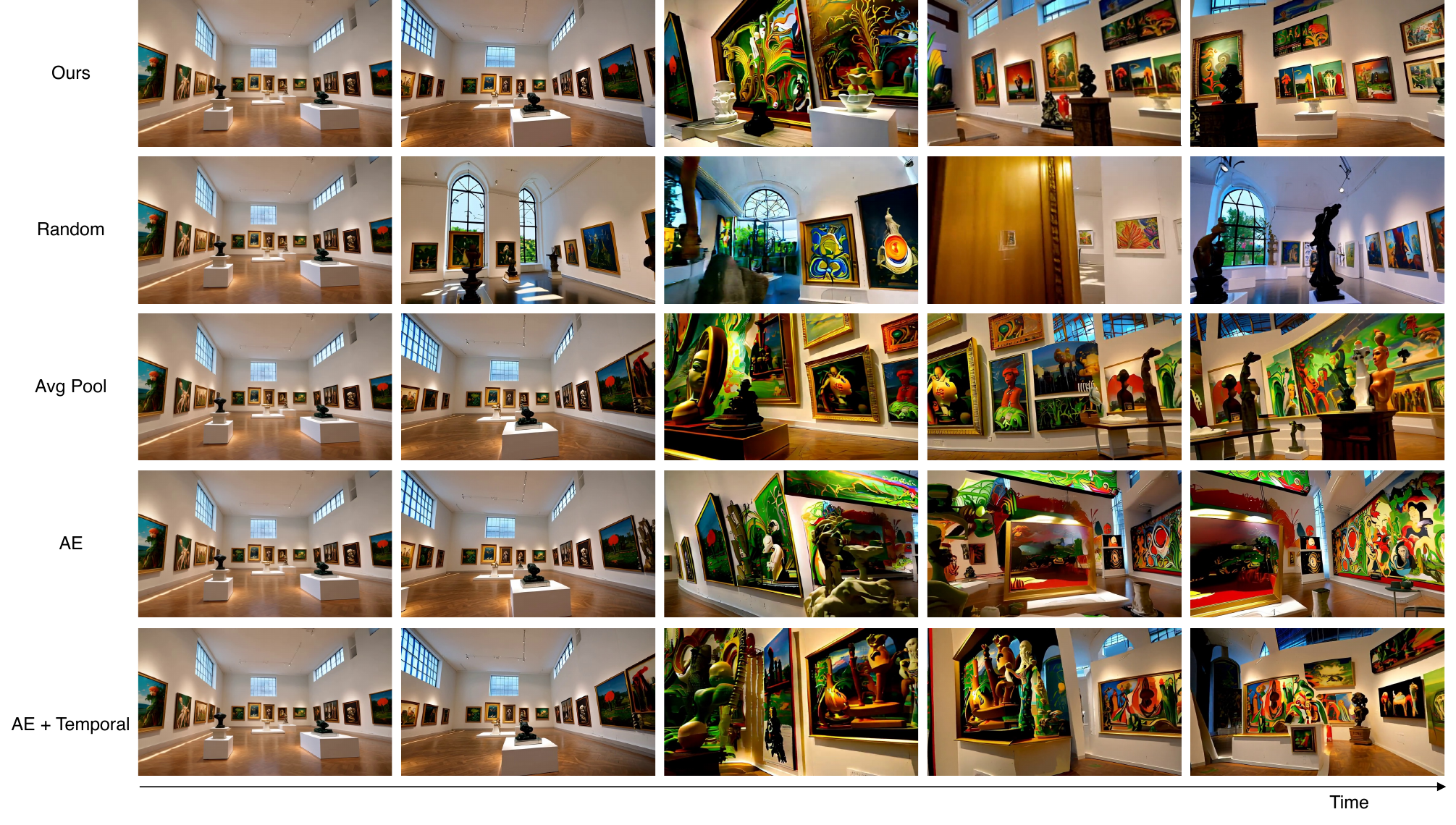}
    \caption{
    Embedding-space comparison.
    The learned LongLive-RAG embedding retrieves more useful historical context than random or hand-crafted alternatives.
    }
    \label{fig:app_ablationMethod}
\end{figure}

\end{document}